\documentclass[letterpaper]{article} % DO NOT CHANGE THIS
\usepackage{aaai2026}  % DO NOT CHANGE THIS
\usepackage{times}  % DO NOT CHANGE THIS
\usepackage{helvet}  % DO NOT CHANGE THIS
\usepackage{courier}  % DO NOT CHANGE THIS
\usepackage[hyphens]{url}  % DO NOT CHANGE THIS
\usepackage{graphicx} % DO NOT CHANGE THIS
\usepackage{natbib}  % DO NOT CHANGE THIS AND DO NOT ADD ANY OPTIONS TO IT
\usepackage{caption} % DO NOT CHANGE THIS AND DO NOT ADD ANY OPTIONS TO IT
\usepackage{algorithm}
\usepackage{algorithmic}
\usepackage{amsmath, amssymb}
\usepackage{enumitem}
\usepackage{booktabs}
\usepackage{newfloat}
\usepackage{listings}
\DeclareCaptionStyle{ruled}{labelfont=normalfont,labelsep=colon,strut=off} % DO NOT CHANGE THIS
\floatstyle{ruled}
\newfloat{listing}{tb}{lst}{}
\floatname{listing}{Listing}
\title{Graph2Video: Leveraging Video Models to Model Dynamic Graph Evolution}
\author{
    Hua Liu\textsuperscript{\rm 1},
    Yanbin Wei\textsuperscript{\rm 1, \rm 2},
    Fei Xing\textsuperscript{\rm 3},
    Tyler Derr\textsuperscript{\rm 4},
    Haoyu Han\textsuperscript{\rm 5},
    Yu Zhang\textsuperscript{\rm 1}\thanks{Corresponding author.}
}
\affiliations{
    \textsuperscript{\rm 1}Southern University of Science and Technology\\
    \textsuperscript{\rm 2}The Hong Kong University of Science and Technology\\
    \textsuperscript{\rm 3}City University of Hong Kong\\
    \textsuperscript{\rm 4}Vanderbilt University\\
    \textsuperscript{\rm 5}Michigan State University\\
    \{liuh5, zhangy7\}@sustech.edu.cn, yanbin.ust@gmail.com, 
    fxing8-c@my.cityu.edu.hk,
    tyler.derr@vanderbilt.edu,
    hanhaoy1@msu.edu
}

\usepackage{bibentry}
\begin{document}

\maketitle

\begin{abstract}
% Dynamic graphs are common in real‑world systems such as social media, recommender systems, and traffic networks. However, most existing methods either discrete time into uniform snapshots or rely on node‑centric memories, which often fail to capture the dynamic evolution between node pairs. We propose Graph2Video, a video‑inspired framework for continuous‑time dynamic graph (CTDG) learning that views the temporal neighborhood of a target link as a sequence of “graph frames”. By stacking local subgraph “patches” over time, we form 3‑D graph cubes that can be processed by existing video foundation models, granting the encoder strong inductive biases for modeling long‑range structural motion. Within Graph2Video we embed a lightweight, link‑specific Dynamic Evolution Pattern Recognizer (DEPR) that explicitly tracks each link’s spatio‑temporal context and complements node memories. Graph2Video unifies the strengths of sequence‑based, memory‑based, and temporal walk‑based approaches in a single architecture: (i) it captures fine‑grained temporal dynamics continuously; (ii) it preserves higher‑order structural patterns through video‑style receptive fields; and (iii) it preserves long-term relational dependencies through compact link memories. Extensive experiments on six public CTDG benchmarks show that Graph2Video consistently surpasses state‑of‑the‑art baselines in link prediction. These results demonstrate that borrowing spatio‑temporal reasoning techniques from computer vision provides a principled and effective avenue for advancing dynamic graph learning.
Dynamic graphs are common in real‑world systems such as social media, recommender systems, and traffic networks. Existing dynamic graph models for link prediction often fall short in capturing the complexity of temporal evolution. 
They tend to overlook fine‑grained variations in temporal interaction order, struggle with dependencies that span long time horizons, and offer limited capability to model pair‑specific relational dynamics.
To address these challenges, we propose \textbf{Graph2Video}, a video‑inspired framework that views the temporal neighborhood of a target link as a sequence of “graph frames”. By stacking temporally ordered subgraph frames into a “graph video”, Graph2Video leverages the inductive biases of video foundation models to capture both fine-grained local variations and long-range temporal dynamics. It generates a link-level embedding that serves as a lightweight and plug-and-play link-centric memory unit. This embedding integrates seamlessly into existing dynamic graph encoders, effectively addressing the limitations of prior approaches.
Extensive experiments on benchmark datasets show that Graph2Video outperforms state‑of‑the‑art baselines on the link prediction task in most cases. The results highlight the potential of borrowing spatio‑temporal modeling techniques from computer vision as a promising and effective approach for advancing dynamic graph learning.

\end{abstract}

\begin{links}
    \link{Code}{https://github.com/hualiu829/Graph2Video}
    % \link{Datasets}{https://aaai.org/example/datasets}
    % \link{Extended version}{https://aaai.org/example/extended-version}
\end{links}

\section{Introduction}
In recent years, dynamic graphs have attracted substantial research interest, owing to the fact that many real-world graph-structured data are inherently dynamic, such as those found in social networks~\cite{shu2017fake,kumar2019predicting,alvarez2021evolutionary,liu2020evolution}, recommender systems~\cite{fan2019graph, zhang2022dynamic, gao2022graph}, and traffic networks~\cite{wang2020traffic, bui2022spatial,sharma2023graph}.
Although graph neural networks (GNNs)~\cite{kipf2016semi, hamilton2017inductive, velivckovic2017graph} have achieved remarkable progress on static graphs~\cite{liu2021elastic,liu2023enhancing} with fixed nodes and edges, they cannot be directly extended to dynamic graphs~\cite{barros2021survey, kazemi2020representation}, where nodes and edges evolve over time. 
In dynamic graphs, complex temporal dependencies and structural changes arise, posing challenges that exceed the modeling capabilities of conventional static GNNs.
As a result, there is a growing demand for dynamic graph learning models that can effectively capture temporal evolution, which enables more precise modeling and analysis of complex, time-evolving systems in practical applications.

Early methods for dynamic graph learning often struggle to capture fine-grained temporal dynamics. 
Most approaches fail to model the precise order and timing of interactions~\cite{ sankar2020dysat,cong2021dynamic,you2022roland}. 
To address these limitations, recent years have witnessed advanced models, including sequence-based methods~\cite{wang2021tcl, cong2023we, yu2023towards, peng2025tidformer}, temporal random walk methods~\cite{wang2021inductive,li2023zebra,lu2024improving}, and memory-based methods~\cite{rossi2020temporal, su2024pres, sheng2024mspipe, ji2024memmap}. 
Although these approaches have advanced the modeling of temporal dynamics, each still exhibits inherent limitations when applied to the \emph{dynamic link prediction} task, which is the primary focus of this work.
Specifically, \textbf{sequence‑based methods} %~\cite{wang2021tcl,  peng2025tidformer} 
build interaction sequences for the two endpoints of a candidate link and encode them using sequential models such as Transformer-based~\cite{vaswani2017attention} or RNN-based~\cite{schuster1997bidirectional} architectures. 
While effective for modeling temporal dependencies, those models~\cite{yu2023towards} are limited by sequence length and may fail to capture long-range dynamics. 
\textbf{Temporal random walk methods} %~\cite{wang2021inductive, souza2022provably, lu2024improving} 
utilize time-respecting paths to explore higher-order neighborhoods and capture contextual information beyond immediate interactions. However, to remain computationally tractable, those methods require truncating the walk length, inevitably discarding portions of the historical context. As a result, they may overlook temporally distant yet informative interactions that could be crucial for predicting future link formation \cite{wang2021inductive}.
\textbf{Memory-based methods} %~\cite{trivedi2019dyrep, kumar2019predicting, rossi2020temporal}
address the limitations of the aforementioned sequence-based and temporal random walk methods by maintaining a persistent memory state for each node, which is continuously updated through recurrent mechanisms to capture long-term dependency information. While effective in preserving node-level historical context, they often lack explicit representations of pairwise relational dynamics~\cite{ji2024memmap}, which are crucial for accurately predicting link formation~\cite{ma2024mixture}.
The above limitations motivate us to explore a novel perspective on dynamic graph learning that can simultaneously capture local fine‑grained variations and global long‑range temporal dependencies, while explicitly modeling pairwise relational dynamics crucial for the dynamic link prediction task.
\paragraph{Graph-as-Video: A New Perspective.}
To overcome the above limitations, drawing inspiration from the success of computer vision and recent progress in vision-enhanced graph learning~\cite{gita, wei2025open}, we introduce a novel \emph{graph-as-video} perspective. Specifically, large-scale video foundation models (e.g., VideoMAE~\cite{tong2022videomae, wang2023videomae}, InternVideo2~\cite{wang2024internvideo2}) have demonstrated the ability to hierarchically disentangle \textit{spatial appearance} from \textit{cross-frame motion}, thereby perceiving both local fine-grained variations and long-range temporal patterns spanning thousands of frames. We observe that the evolving node-edge topology in dynamic graphs is highly isomorphic to the spatio-temporal transformations of pixel distributions in video sequences: a single frame corresponds to a \textbf{structural snapshot} of the dynamic graph at a specific time, while cross-frame motions align with the \textbf{evolutionary trajectories} of node interactions. Thus, by serializing a candidate link and its spatio-temporal neighborhood into a “graph video”, we can effectively inherit the inductive biases of video models for complex temporal regularities at virtually no additional cost.
Building on this perspective, we propose \textbf{Graph2Video}, a lightweight framework that implements the graph-as-video idea for dynamic graph learning, which can be seamlessly integrated into existing methods.

Graph2Video addresses three major challenges in the dynamic link prediction task. \textbf{First}, by rendering temporally ordered subgraph frames, Graph2Video preserves \emph{fine-grained motion cues}: for example, the precise order in which a triangle closes becomes observable as edge appearances across consecutive frames.  
\textbf{Second}, by feeding the entire graph video into a video model optimized for capturing \emph{long-range dependencies}, the model naturally attends to motifs and interaction patterns whose influence may extend across thousands of events.  
\textbf{Third}, the resulting link-level spatio-temporal embedding acts as a \textbf{link-centric memory unit}, explicitly encoding the relational dynamics of the target link. This complements memory-based methods, which mainly focus on node-centric memories and often overlook pairwise structures critical for link prediction. 

To instantiate Graph2Video, at each time step, we capture the local neighborhood surrounding the two endpoints of a candidate link and render it as a structural frame. In those frames, endpoints are explicitly highlighted, together with the timing of edge formations and the connectivity among their historical neighbors. Arranging these frames chronologically yields a compact “graph video” clip that depicts how the substructure of the candidate link evolves over time.
Feeding this video into a frozen video model produces a rich link-level embedding that can be seamlessly integrated into existing dynamic graph encoders. This significantly enhances performance on the dynamic link prediction task without increasing sequence length or random-walk depth, while also addressing the limitations of node-centric memory paradigms in modeling pairwise relational dynamics.
% Prior studies in complex and social network~\cite{zhou2009predicting, bianconi2014triadic, weng2013role} have shown that the probability of link formation is strongly influenced by the \textbf{local sub-structural similarity} of its endpoints. For example, two nodes are more likely to connect in the future if they participate in a triadic closure or share a dense neighborhood within a similar time span. 
% % By visualizing such structural “actions” and feeding them into a frozen video backbone, 
% By visualizing these evolving structural patterns as a sequence of coherent “graph motions” and feeding the resulting graph‑video into a frozen video backbone,
% % By visualizing such evolving structures as coherent graph motions and encoding them with a frozen video backbone，
% we obtain a \textbf{link-level spatio-temporal embedding} that encodes both short-term interaction patterns and long-range co-evolution across events. This embedding serves as a \textbf{link-centric memory unit}, which is fused into the representations of the corresponding nodes within existing CTDG encoders.
% In doing so, we significantly enhance performance on dynamic link prediction, without increasing sequence length or random-walk depth, while addressing the deficiency of node-centric memory paradigms in capturing pairwise relational dynamics.

Our main contributions are summarized as follows.
\begin{itemize}[leftmargin=1.6em]
    \item \textbf{Graph2Video Framework.}  
          We propose Graph2Video, a novel and model‑agnostic framework that reformulates link‑centered dynamic graph neighborhoods as “graph videos”, enabling principled modeling of spatio‑temporal dynamics. The framework is lightweight and plug‑and‑play, making it applicable to a wide range of dynamic graph models without architectural modifications.  
    \item \textbf{Link‑Centric Memory Injection.}  
          We introduce a lightweight mechanism that augments dynamic graph encoders with link‑level spatio‑temporal embeddings, explicitly capturing pairwise relational dynamics while simultaneously preserving fine‑grained temporal cues and long‑range dependencies.  
    \item \textbf{Empirical Gains.}  
          Extensive experiments on five public benchmarks show that Graph2Video consistently enhances diverse backbones and achieves state‑of‑the‑art performance in dynamic link prediction tasks, demonstrating both its effectiveness and generality.  
\end{itemize}

\section{Related Work}
\subsection{Dynamic Graph Learning}
Early efforts in dynamic graph learning struggled to model fine-grained temporal dynamics. Approaches based on GNNs~\cite{xu2020inductive, wang2021tcl, yu2023towards} or recurrent architectures~\cite{schuster1997bidirectional, vaswani2017attention} captured overall structural and sequential patterns but often failed to preserve the exact order and timing of interaction information~\cite{sankar2020dysat, cong2021dynamic, you2022roland}.

Recent progress falls into three main research lines.
Sequence-based methods build chronologically ordered interaction sequences for link endpoints and encode them with RNN-based or Transformer-based architectures to capture both short-term variations and long-range dependencies. Although models such as TCL~\cite{wang2021tcl}, DyGFormer~\cite{yu2023towards}, and TIDFormer~\cite{peng2025tidformer} enhance Transformers with dedicated temporal encodings and interaction-aware mechanisms, the quadratic cost of self-attention necessitates short sequence lengths, limiting their ability to capture very long‑range dynamics.

Temporal random-walk methods recover higher-order historical context by sampling time-consistent paths. 
CAWN~\cite{wang2021inductive} aggregates causal anonymous walks; PINT~\cite{souza2022provably} performs injective temporal message passing with walk-count positional features; TPNet~\cite{lu2024improving} projects temporal walk matrices via random feature propagation.
Despite these advances, all three must truncate walk lengths or branching factor for tractability, inevitably discarding temporally distant yet informative interactions that could reveal future links.

Memory-based methods maintain a persistent trainable state for each node, updated after every event to accumulate long-term dependency information. Early works such as DyRep~\cite{trivedi2019dyrep} and JODIE~\cite{kumar2019predicting} use recurrent updates, while more recent models like TGN~\cite{rossi2020temporal} and MemMap~\cite{ji2024memmap} introduce richer aggregation and hierarchical memory. Although effective, these approaches remain node-centric and lack explicit modeling of pairwise relational dynamics that are crucial for link prediction.

% Despite their success, all three paradigms face fundamental constraints: sequence models are limited by sequence length, random-walk models by sampling truncation, and memory-based models by their lack of explicit pairwise evolution modeling. These challenges motivate new perspectives capable of jointly capturing local fine-grained variations and global long-range dependencies in dynamic graphs.
While these methods have achieved notable success, existing paradigms still face fundamental limitations: Sequence models are constrained by limited sequence lengths, random walk models suffer from sampling truncation, and memory‑based models overlook the explicit evolution of pairwise relationships. These limitations motivate the need for new perspectives on dynamic graph learning for link prediction that can jointly capture local fine-grained variations and global long-range dependencies.

\subsubsection{Video Foundation Model}
Early work relied on hand-crafted features or optical flow, followed by 3D CNNs such as I3D and C3D for capturing local spatiotemporal patterns, and later by Transformer-based architectures such as TimeSformer~\cite{gberta_2021_ICML} and MViT~\cite{fan2021multiscale} for global modeling.
More recently, large-scale video foundation models have significantly improved cross-task transferability and generalization. Among them, the VideoMAE series represents the masked auto-encoding paradigm. VideoMAE~\cite{tong2022videomae} randomly masks most spatiotemporal patches and reconstructs the video cube to learn disentangled appearance–motion features. Its successor, VideoMAE V2~\cite{wang2023videomae}, scales up both the model and the training pipeline, achieving strong performance across benchmarks. Another influential line is the SAM series, such as SAM2~\cite{ravi2024sam}, which extends prompt-driven segmentation to videos with a streaming memory mechanism, enabling high-precision multi-frame tracking and segmentation.

As model and data scales grow, the spatiotemporal inductive biases of video foundation models offer natural advantages for dynamic graph modeling. Motivated by this, we propose Graph2Video, which converts dynamic graphs into temporally ordered “graph videos”. At each timestamp, we extract the local subgraph, divide it into topological patches, and concatenate these frames to form a temporal graph representation suitable for video models. This enables direct use of masked self-supervised strategies and hierarchical spatiotemporal encoders. Compared with static graph encoders, Graph2Video yields time-aware link representations and naturally incorporates video-inspired memory mechanisms to capture long-range dependencies, providing a new perspective for dynamic link prediction.

% As model and data scales continue to grow, the spatiotemporal inductive biases of video foundation models exhibit natural advantages for dynamic graph modeling. Motivated by this insight, we propose Graph2Video, which reconstructs dynamic graphs as temporally ordered “graph videos.” Specifically, at each timestamp we extract the local subgraph, partition it into topological patches, and then concatenate these structured frames in chronological order to form a temporal graph representation suitable for video models. This design enables the direct use of masked self-supervised learning strategies and hierarchical spatiotemporal encoders from models such as VideoMAE V2. Compared with conventional static graph encoders, Graph2Video produces time-aware link representations and naturally incorporates video-inspired memory mechanisms to capture long-range dependencies, offering a new perspective for link prediction in dynamic graphs.
\section{Preliminary}
\paragraph{Dynamic Graph.} Given a node set $\mathcal{V}$ and an edge set $\mathcal{E}$, a dynamic graph $\mathcal{G}$ can be represented as a sequence of chronologically ordered interactions denoted by $\mathcal{G} = \{ (u_1, v_1, t_1), (u_2, v_2, t_2), \cdots, (u_k, v_k, t_k)\}$ with $0 \leq t_1 \leq t_2 \leq \cdots\leq t_k$, where $u_i, v_i \in \mathcal{V}$ are the source and destination nodes of the newly formed link at timestamp $t_i$, respectively. Each node $v \in \mathcal{V}$ can be associated with node feature $\boldsymbol{x}_v \in \mathbb{R}^{d_V}$, and each interaction $(u, v, t) \in \mathcal{E}_t$ has link feature $\boldsymbol{e}_{u,v}(t) \in \mathbb{R}^{d_E}$. Here, $d_V$ and $d_E$ denote the dimensions of the node feature and the link feature, respectively. For non-attributed graphs, we simply set the node feature and link feature to zero vectors, i.e., $\boldsymbol{x}_u = \boldsymbol{0}$ and $\boldsymbol{e}_{u,v}(t) = \boldsymbol{0}$. 

\paragraph{Dynamic Graph Link Prediction.} 
Given two nodes $u$ and $v$, a timestamp $t$, and the historical interactions prior to $t$ (i.e., $ \mathcal{H}_{<t} = \{(u', v', t') \mid t' < t\}$), the goal of dynamic graph link prediction is to learn a function  
\[
f: (u, v, t, \mathcal{H}_{<t}) \rightarrow \{0, 1\}
\]  
that predicts whether a link between $u$ and $v$ will occur at time $t$, based on the temporal and structural context encoded in the historical interaction set $\mathcal{H}_{<t}$.

% \textbf{Definition 3} (K-hop Subgraph). We use the notation $\mathcal{G}(t) = (\mathcal{V}(t), \mathcal{E}(t))$ to denote the graph snapshot at $t$, where $\mathcal{E}(t)$ includes all the interactions that happen before $t$ and $\mathcal{V}(t)$ includes all the nodes appear in $\mathcal{E}(t)$. Besides, we defined the $k$-hop subgraph of node $u$ as $\mathcal{G}_u^k(t) = (\mathcal{V}_u^k(t), \mathcal{E}_u^k(t))$, where $\mathcal{V}_u^k(t) \subset \mathcal{N}(t)$ is the set of nodes whose shortest path distance to $u$ is less than $k$ on $\mathcal{G}(t)$ and $\mathcal{E}_u^k(t) \subset \mathcal{E}(t)$ is the set of interactions between $\mathcal{V}_u^k(t)$.

% \subsection{Definition 4. Temporal Walk.}
% A $k$-step temporal walk $W$ on $\mathcal{G}(t)$ is a sequence of node-time pair with \textcolor{yellow}{decreased temporal order}~\cite{xu2020inductive}, which can be denoted as $W = [(w_0, t_0), \cdots, (w_k, t_k)]$ with $t = t_0 > t_1 > \cdots > t_k$ and $(\{w_i, w_{i+1}\}, t_{i+1})$ is an edge on $\mathcal{G}(t)$ for $0 \le i \le k-1$. Figure 2 shows a visual illustration of a temporal walk. Here, we stipulate the first timestamp $t_0$ as the current time $t$ to avoid undefinedness of $t_0$. We use the notation $\mathcal{M}_{u,v}^k(t)$ to denote the set of all $k$-step temporal walks from $u$ to $v$ on $\mathcal{G}(t)$. Specially, $\mathcal{M}_{u,v}^0(t) = \{[(u, t)]\}$ if $u = w$ and $\mathcal{M}_{u,v}^0(t) = \emptyset$ otherwise. When there is no ambiguity, we replace $\mathcal{M}_{u,v}^k(t)$ with $\mathcal{M}_{u,v}^k$.

%
\begin{figure*}[t]
\centering
\includegraphics[width=0.95\linewidth]{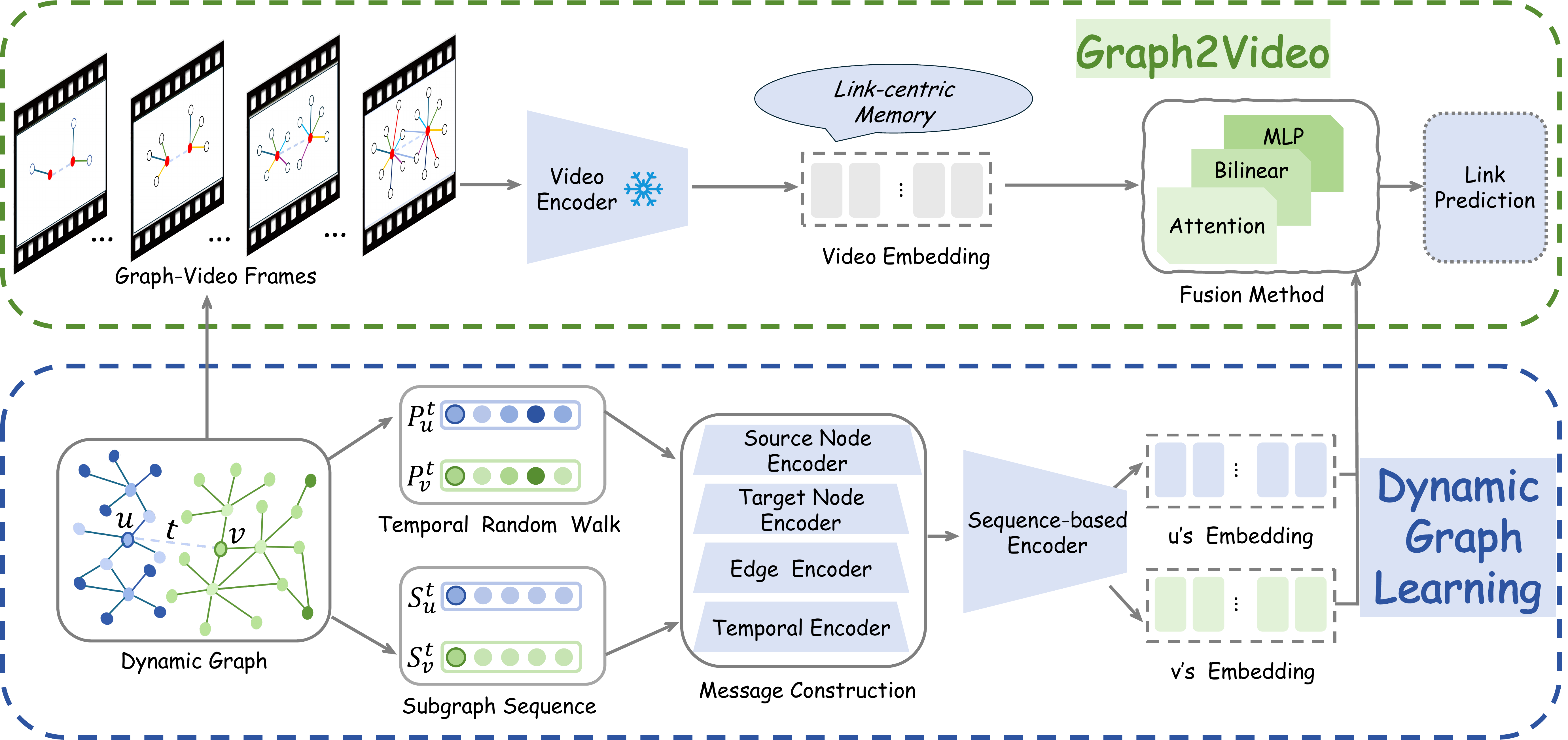} 
\caption{The framework of Graph2Video.}
\label{fig: framework}
\end{figure*}

\section{Methodology}
We introduce Graph2Video, a lightweight framework that injects vision‑aware temporal patterns into existing dynamic graph link predictors. As shown in Figure~\ref{fig: framework}, Graph2Video comprises four modular components:  
(1) \textit{Graph-Video Construction},  
(2) \textit{Graph-Motion Extraction},  
(3) \textit{Plug-and-Play Dynamic Graph Encoder}, and  
(4) \textit{Adaptive Feature Fusion}.  
We now describe each component in detail.

\subsection{Graph-Video Construction}
% \begin{figure}[t]
% \small
% \centering
% \includegraphics[width=0.45\textwidth]{AnonymousSubmission/LaTeX/gennerated video.png} % Reduce the figure size so that it is slightly narrower than the column.
% \caption{An illustrative example of graph-video construction over temporal slices.}
% \label{fig:graphvideo}
% \end{figure}
% \label{sec:scope_decoupled}
We first convert the temporal evolution of a target link \((u,v)\) into a fixed-length “graph video”. The key to capturing link dynamics lies in modeling \emph{two complementary scopes}: (i) the \textbf{spatial scope}, i.e., which part of the graph is observed around the query link, and (ii) the \textbf{temporal scope}, i.e., how the observed structure evolves over time. We therefore construct a \emph{scope-decoupled subgraph video} that isolates spatial and temporal contexts before feeding them into the frozen video backbone.

\paragraph{Temporal Scope (uniform slicing).}
Given a target link \((u,v)\) and a prediction time \(t^{*}\),
let \(t_{0}\) be the timestamp of the earliest event incident on \(u\) or \(v\).
We define the temporal window as the interval $[t_0, t^*]$, which captures all historical patterns that
may affect the formation of $(u,v)$. This interval is uniformly divided into $F$ equal-length segments, and graph snapshots at the end of each segment yield an ordered frame set:
$\{\,\mathcal{G}^{(u,v)}_{t_1}, \dots, \mathcal{G}^{(u,v)}_{t_F}\,\}$, with $t_1 < \cdots < t_F = t^*$.

% \hy{Is this DTDG? equal time interval?}
% \(\{\mathcal{G}^{(u,v)}_{t_1},\dots,\mathcal{G}^{(u,v)}_{t_F}\}\) with \(t_F=t^*\).  

% \paragraph{Spatial Scope (local structural context).}
% At each time point $t_i$, we construct a local subgraph centered on the target link $(u,v)$ by inducing the 1-hop temporal neighborhood of each endpoint. For each node $x \in \{u,v\}$, we define $\mathcal{V}_x(t_i) = \mathrm{Top}^k\left(N_1(x,t_i)\right)$, where $N_1(x,t_i)$ is the multiset of nodes that interacted with $x$ prior to $t_i$, and $\mathrm{Top}^k(\cdot)$ selects the $k$ most recent neighbors ranked by interaction time. The full node set of the subgraph is then given by $\mathcal{V}^{1}_{u,v}(t_i) = \{u,v\} \cup \mathcal{V}_u(t_i) \cup \mathcal{V}_v(t_i)$, and the corresponding induced subgraph is $\mathcal{G}^{(u,v)}_{t_i} = \left( \mathcal{V}^{1}_{u,v}(t_i),\, \mathcal{E}^{1}_{u,v}(t_i) \right)$, where $\mathcal{E}^{1}_{u,v}(t_i)$ includes all edges among nodes in $\mathcal{V}^{1}_{u,v}(t_i)$ that exist at time $t_i$. This construction captures a temporally aligned, layout-invariant structural context around the link, facilitating consistent modeling across dynamic snapshots.

\paragraph{Spatial Scope (local structural context).}
At each time point $t_i$, we construct a local subgraph centered on the target link $(u,v)$ by inducing the $k$-hop temporal neighborhood of each endpoint. For each node $x \in \{u,v\}$, we define $\mathcal{V}^k_x(t_i) = \mathrm{Top}^s\left(N_k(x,t_i)\right)$, 
% where $N_k(x,t_i)$ is the multiset of nodes that interacted with $x$ prior to $t_i$, 
where $N_k(x,t_i)$ denotes the multiset of nodes within $k$-hop temporal neighborhoods 
that interacted with $x$ prior to $t_i$,
% where $N_k(x,t_i)$ denotes the multiset of nodes reachable from $x$ through up to $k$ 
% temporal hops, capturing all interactions that occurred prior to $t_i$,
and $\mathrm{Top}^s(\cdot)$ selects the $s$ most recent neighbors ranked by interaction time. The full node set of the subgraph is then given by $\mathcal{V}^k_{u,v}(t_i) = \{u,v\} \cup \mathcal{V}^k_u(t_i) \cup \mathcal{V}^k_v(t_i)$, and the corresponding induced subgraph is $\mathcal{G}^{(u,v)}_{t_i} = \left( \mathcal{V}^k_{u,v}(t_i),\, \mathcal{E}^k_{u,v}(t_i) \right)$, where $\mathcal{E}^k_{u,v}(t_i)$ includes all edges among nodes in $\mathcal{V}^k_{u,v}(t_i)$ at time $t_i$. 

% \hy{Can you say we use the K-hop neighbors? In experiments, we select k=1, it is weird that only use 1-hop neibors to capture the global dependency as you said before. }

\paragraph{Qualitative Subgraph Visualization}
\label{sec:viz}
To qualitatively verify that our scope-decoupled construction retains meaningful temporal dynamics, we visualize each subgraph frame $\mathcal{G}^{(u,v)}_{t_i}$ using automated graph visualization tools, including \texttt{Graphviz}~\cite{gansner2000open} and \texttt{Matplotlib}~\cite{tosi2009matplotlib}. This visualization also provides an intuitive view of how local structures evolve over time.
To ensure consistent layouts and avoid introducing noise caused by layout variation, we carefully standardize every aspect of the visualization canvas, including the graph layout algorithm, node colors, and node shapes. Specifically, 
the two endpoints are pinned to predefined anchor positions; visual cues are added by coloring endpoints in red, arranging their respective $s$ most recent neighbors in separate clusters, color-coding edges by interaction time, and inserting dummy placeholder nodes for absent neighbors so that node ordering and layout remain stable across frames. Each resulting high-resolution PNG is converted to RGB format and resized to a fixed spatial resolution $(H,W)$, yielding a video tensor $\boldsymbol{X}^{(i)} \in \mathbb{R}^{3\times H\times W}$.  
Stacking the $F$ frame tensors along the temporal axis produces the final video input $\boldsymbol{X} \in \mathbb{R}^{F\times3\times H\times W}$, which is fed \emph{unchanged} into the frozen video backbone. Therefore, this visualization pipeline offers interpretability without influencing training or inference and yields a temporally ordered tensor stack that we interpret as a “graph motion sequence”, where the appearance or disappearance of edges around $(u,v)$ constitutes the motion. This construction captures a temporally aligned, layout-invariant structural context around the link, facilitating consistent modeling across dynamic snapshots. As shown in Figure~\ref{fig: framework}, the top-left panel labeled “Graph-Video Frames” illustrates the stacked frames. Each frame depicts the 1-hop local subgraph centered on the target link $(u,v)$ across different temporal slices.

% An illustrative example of the constructed graph video is shown in Figure~\ref{fig: framework}, where each frame corresponds to the 1-hop local subgraph surrounding the target link $(u,v)$ at different temporal slices.
\subsection{Graph-Motion Extraction}
Having constructed scope-decoupled graph videos, the next step is to extract expressive spatio-temporal features that capture the structural dynamics around a candidate link. 
To this end, we leverage pre-trained video backbones, which are well-suited for recognizing both fine-grained motions and long-range temporal dependencies. 
However, directly training or fine-tuning such large-scale models on graph videos is often infeasible, motivating a frozen but expressive design.

\label{sec:videoFeature}

\paragraph{Graph–Motion Feature Extractor.} Each scope\-decoupled subgraph video is represented as a video tensor $\boldsymbol{X}$, which we feed into a \emph{pre‑trained frozen} video backbone $\phi^{\text{video}}$. The backbone produces an embedding 
\[\boldsymbol{f}^{\text{video}}_{uv}= \phi^{\text{video}}(\boldsymbol{X})\in \mathbb{R}^{d_{\text{vid}}},\]
where $d_{\text{vid}}$ is the output dimensionality of the chosen backbone. In principle, $\phi^{\text{video}}$ can be instantiated by any off‑the‑shelf video model that performs spatio-temporal attention (e.g., TimeSformer~\cite{gberta_2021_ICML}, ViViT~\cite{arnab2021vivit}). In our implementation, we adopt VideoMAE V2~\cite{wang2023videomae}, a masked autoencoding Transformer that reconstructs sparsely sampled video cubes; substituting another backbone requires no changes to the rest of the pipeline. Because the video encoder is frozen, without any additional training, the backbone outputs a spatio‑temporal embedding $\boldsymbol{f}^{\text{video}}_{uv}$ encoding high‑level “structural motions” such as triadic closures or neighbor dispersal patterns around the link $(u,v)$. We regard this embedding as a \textbf{link\textendash centric memory unit}: it preserves the spatio–temporal history of the pair $(u,v)$, augments the node–centric states maintained by conventional dynamic graph encoders, and supplies the fine–grained and long–range relational cues that those encoders lack.

\paragraph{Frozen but Expressive.}
Fine-tuning such powerful video
models on graph videos is both computationally expensive and
prone to overfitting, due to the lack of rich low-level textures in
graph-structured inputs.
Instead, we freeze the backbone $\phi^{\text{video}}:\mathbb{R}^{F\times 3\times H\times W}\!\to\!\mathbb{R}^{d_{\text{vid}}}$ and use it solely as a feature generator. This design brings three notable benefits. First, it improves \textbf{efficiency}, as video embeddings are computed once and cached, thereby reducing both training time and memory consumption. Second, it enhances \textbf{regularization}, since freezing prevents catastrophic forgetting and preserves the motion-sensitive filters learned from large-scale video corpora. Third, it provides strong \textbf{domain robustness}, as shown in our experiments (Section~\ref{lab: experiment}), where frozen video features transfer effectively despite the modality gap, thereby validating the “graph-as-video” hypothesis.

\subsection{Plug‑and‑Play Dynamic Graph Encoder}
In this work, we focus on learning expressive representations for dynamic link prediction. 
A key challenge of this task is capturing \emph{long‑range temporal dependencies}, 
which arise not only from extended interaction histories but also from higher‑order structural contexts. 
To tackle this challenge, we adopt two representative dynamic graph encoders, DyGFormer and TPNet, that approach long‑term dependencies from complementary perspectives.
  \textbf{DyGFormer}~\cite{yu2023towards} is a \emph{sequence‑based} encoder that applies time‑aware self‑attention over chronologically ordered event histories, 
capturing long‑range dependencies through temporal interaction patterns.  
\textbf{TPNet}~\cite{lu2024improving} is a \emph{temporal random walk‑based} encoder that models higher‑order structural dependencies by projecting temporal walk matrices 
via random feature propagation, avoiding explicit path sampling while maintaining efficiency.  

To demonstrate that Graph2Video integrates seamlessly with dynamic graph encoders and boosts their ability to capture long-range dependencies without architectural modifications or additional supervision, we retain the original architectures and hyperparameter settings of the encoders.

\subsection{Adaptive Feature Fusion}
\begin{table*}[!ht]
\centering
\small
% \footnotesize
\setlength{\tabcolsep}{1mm}{
\begin{tabular}{l|ccccccccc}
\toprule
NSS & Datasets & TGN & CAWN & TCL & GraphMixer & DyGFormer & DyGFormer+ & TPNet & TPNet+ \\
\midrule
 & Reddit & $98.63_{ \pm 0.06}$ & $99.11_{ \pm 0.01}$ & $97.53_{ \pm 0.02}$ & $97.31_{ \pm 0.01}$ & $99.22_{ \pm 0.01}$ &$99.23_{ \pm 0.01}$  & $\underline{99.27_{ \pm 0.00}}$ & \textbf{99.32}$_{ \pm 0.01}$ \\
 & MOOC & $89.15_{ \pm 1.60}$ & $80.15_{ \pm 0.25}$ & $82.38_{ \pm 0.24}$ & $82.78_{ \pm 0.15}$ & $87.52_{ \pm 0.49}$ & $87.33_{ \pm 0.43}$ & \textbf{96.39}$_{ \pm 0.09}$ & \underline{$96.38_{ \pm 0.03}$} \\
rnd & Enron & $86.53_{ \pm 1.11}$ & $89.56_{ \pm 0.09}$ & $79.70_{ \pm 0.71}$ & $82.25_{ \pm 1.02}$ & $92.47_{ \pm 0.12}$ & $92.32_{ \pm 0.18}$ & \underline{$92.90_{ \pm 0.17}$} & \textbf{93.01}$_{ \pm 0.10}$ \\
 & UCI & $92.34_{ \pm 1.04}$ & $95.18_{ \pm 0.06}$ & $89.57_{ \pm 1.63}$ & $93.25_{ \pm 0.57}$ & $95.79_{ \pm 0.17}$ & $96.24_{ \pm 0.04}$ & \underline{$97.35_{ \pm 0.04}$} & \textbf{97.37}$_{ \pm 0.10}$ \\
 & Can. Parl. & $70.28_{ \pm 2.34}$ & $69.82_{ \pm 2.34}$ & $68.67_{ \pm 2.67}$ & $77.04_{ \pm 0.96}$ & $97.36_{ \pm 0.45}$ & \underline{$97.60_{ \pm 0.40}$} & $90.28_{ \pm 0.37}$ & \textbf{98.70}$_{ \pm 0.13}$ \\
 \cmidrule{2-10}
 & Avg. rank & 5.60 & 6.00 & 7.60 & 6.40 & 3.60 & 3.40 & 2.20 & 1.20 \\
\midrule
 & Reddit & $81.22_{ \pm 0.61}$ & $80.82_{ \pm 0.45}$ & $77.14_{ \pm 0.16}$ & $78.44_{ \pm 0.18}$ & $ \underline{81.57_{ \pm 0.67}}$ & $81.41_{ \pm 1.6}$ & $81.02_{ \pm 1.31}$ & \textbf{82.19}$_{ \pm 1.40}$ \\
 & MOOC & $87.06_{ \pm 1.93}$ & $74.05_{ \pm 0.95}$ & $77.06_{ \pm 0.41}$ & $77.77_{ \pm 0.92}$ & $85.85_{ \pm 0.66}$ & $87.56_{ \pm 0.89}$ & \underline{$92.69_{ \pm 0.95}$} & \textbf{92.94}$_{ \pm 0.17}$ \\
hist & Enron & $73.91_{ \pm 1.76}$ & $64.73_{ \pm 0.36}$ & $70.66_{ \pm 0.39}$ & $77.98_{ \pm 0.92}$ & $75.63_{ \pm 0.73}$ & $77.46_{ \pm 0.72}$ & \underline{$80.79_{ \pm 1.68}$} & \textbf{82.52}$_{ \pm 0.84}$ \\
 & UCI & $80.43_{ \pm 2.12}$ & $65.30_{ \pm 0.43}$ & $80.25_{ \pm 2.74}$ & $84.11_{ \pm 1.35}$ & $82.17_{ \pm 0.82}$ & $82.55_{ \pm 1.13}$ & \underline{$86.34_{ \pm 0.80}$} & \textbf{86.57}$_{ \pm 0.79}$ \\
 & Can. Parl. & $68.42_{ \pm 3.07}$ & $66.53_{ \pm 2.77}$ & $65.93_{ \pm 3.00}$ & $74.34_{ \pm 0.87}$ & $97.00_{ \pm 0.31}$ & \underline{$97.46_{ \pm 0.33}$} & $86.61_{ \pm 3.47}$ & \textbf{97.66}$_{ \pm 0.31}$ \\
 \cmidrule{2-10}
 & Avg. rank & 5.20 & 7.40 & 7.40 & 4.80 & 4.00 & 3.20 & 3.00 & 1.00  \\
\midrule
 & Reddit & $88.10_{ \pm 0.24}$ & \textbf{91.67}$_{ \pm 0.24}$ & $87.45_{ \pm 0.29}$ & $85.26_{ \pm 0.11}$ & $91.11_{ \pm 0.40}$ & $\underline{91.41_{ \pm 0.71}}$ & $88.19_{ \pm 0.33}$ & $88.62_{ \pm 1.10}$ \\
 & MOOC & $77.50_{ \pm 2.91}$ & $73.51_{ \pm 0.94}$ & $74.65_{ \pm 0.54}$ & $74.27_{ \pm 0.92}$ & $81.24_{ \pm 0.69}$ & $82.24_{ \pm 1.20}$ & \underline{$88.18_{ \pm 0.97}$} & \textbf{88.22}$_{ \pm 0.69}$ \\
ind & Enron & $70.89_{ \pm 2.72}$ & $75.15_{ \pm 0.58}$ & $71.29_{ \pm 0.32}$ & $75.01_{ \pm 0.19}$ & \underline{$77.41_{ \pm 0.89}$} & \textbf{78.60}$_{ \pm 0.89}$ & $75.36_{ \pm 1.81}$ & 77.03$_{ \pm 0.86}$ \\
 & UCI & $70.94_{ \pm 0.71}$ & $64.61_{ \pm 0.48}$ & $76.01_{ \pm 1.11}$ & $\textbf{80.10}_{ \pm 0.51}$ & $72.25_{ \pm 1.71}$ & $72.35_{ \pm 1.78}$ & $77.26_{ \pm 1.57}$ & \underline{78.48$_{ \pm 1.27}$} \\
 & Can. Parl. & $65.34_{ \pm 2.87}$ & $67.75_{ \pm 1.00}$ & $65.85_{ \pm 1.75}$ & $69.48_{ \pm 0.63}$ & $95.44_{ \pm 0.57}$ & \underline{$96.39_{ \pm 0.54}$} & $85.59_{ \pm 3.08}$ & \textbf{98.76}$_{ \pm 0.11}$ \\
 \cmidrule{2-10}
 & Avg. rank & 6.80 & 5.60 & 6.20 & 5.40 & 3.60 & 2.60 & 3.60 & 2.20 \\
\bottomrule
\end{tabular}}
\caption{AP (\%) for \textit{transductive} dynamic link prediction on real-world datasets with three sampling strategies (NSS).}
\label{tab1}
\end{table*}

\textbf{Dynamic Link Prediction.} Given two nodes $u,v$ and time $t$, what is the probability that a new interaction will occur (or reoccur) between them?
After obtaining the visual subgraph embedding $\boldsymbol{f}^{\text{video}}_{uv}$, we fuse this \emph{vision-aware cue} with the topology-aware node states $\boldsymbol{h}_u(t),\boldsymbol{h}_v(t)$ ($\boldsymbol{h}_u,\boldsymbol{h}_v$ for brevity) produced by the dynamic graph encoder, and feed the fused embeddings to a standard link decoder to estimate this probability denoted as $\widehat{p}_{uv}(t)$.

Let $\boldsymbol{h}_x\in\mathbb{R}^{d}$ with $x \in \{u,v\}$ be the node embedding and $\boldsymbol{f}^{\text{video}}_{uv}\in\mathbb{R}^{d_{\text{vid}}}$ be the visual feature. We project every modality into a common $d$-dimensional space:
\[
\tilde{\boldsymbol h}_u = \boldsymbol{W}_u \boldsymbol{h}_u,\quad
\tilde{\boldsymbol h}_v = \boldsymbol{W}_v \boldsymbol{h}_v,\quad
\tilde{\boldsymbol f}   = \boldsymbol{W}_f \boldsymbol{f}^{\text{video}}_{uv}.
\]
We provide three plug-and-play fusion modules, \emph{attention-guided}, \emph{bilinear}, and \emph{MLP}, that produce vision-enhanced node embeddings
$\widehat{\boldsymbol y}_x\in\mathbb{R}^{d}$, $x \in \{u,v\}$ without altering the underlying dynamic graph backbone. 
\begin{itemize}[leftmargin=1.4em,topsep=2pt,itemsep=2pt]
    \item \textbf{Attention-guided.}\;  
  Cross-modal multi-head attention aligns $\tilde{\boldsymbol h}_x$ with $\tilde{\boldsymbol f}$, then mixes them via a learnable gate $\alpha=\psi(\theta)$:
  \[
  \boldsymbol q_x=\mathrm{FFN}\!\bigl(\mathrm{Attn}(\tilde{\boldsymbol h}_x,\tilde{\boldsymbol f},\tilde{\boldsymbol f})\bigr),\;
  \widehat{\boldsymbol y}_x=(1-\alpha)\tilde{\boldsymbol h}_x+\alpha\,\boldsymbol q_x.
  \]
  \item \textbf{Bilinear.}\;  
  A tensor $\boldsymbol W\in\mathbb{R}^{d\times d\times d}$ captures second-order interactions:
  \[
  \widehat{\boldsymbol y}_x^{(k)}=\tilde{\boldsymbol h}_x^{\top}\boldsymbol W_k\tilde{\boldsymbol f}+ \boldsymbol b_k,
  \]
  where $\boldsymbol{W}_k \in \mathbb{R}^{d \times d}$ is the $k$-th slice and $\boldsymbol b_k$ is a bias term.
  \item \textbf{MLP.}\;  
  A lightweight two-layer perceptron on the concatenation $[\tilde{\boldsymbol h}_x\parallel \tilde{\boldsymbol f}]$:
  \[
  \widehat{\boldsymbol y}_x
        =\sigma \!\left(\boldsymbol W_2 \cdot \sigma\!\left(\boldsymbol W_1[\tilde{\boldsymbol h}_x\parallel \tilde{\boldsymbol f}]
                         +\boldsymbol b_1\right)+\boldsymbol b_2\right).
  \]
\end{itemize}

% \noindent\textbf{Attention-guided.}\;  
%   Cross-modal multi-head attention aligns $\tilde{\boldsymbol h}_x$ with $\tilde{\boldsymbol f}$, then mixes them via a learnable gate $\alpha=\sigma(\theta)$:
%   \[
%   \boldsymbol p_x=\mathrm{FFN}\!\bigl(\mathrm{Attn}(\tilde{\boldsymbol h}_x,\tilde{\boldsymbol f},\tilde{\boldsymbol f})\bigr),\;
%   \widehat{\boldsymbol y}_x=(1-\alpha)\tilde{\boldsymbol h}_x+\alpha\,\boldsymbol p_x.
%   \]
% \noindent\textbf{Bilinear.}\;  
%   A tensor $\boldsymbol W\in\mathbb{R}^{d\times d\times d}$ captures second-order interactions:
%   \[
%   \widehat{\boldsymbol y}_x^{(k)}=\tilde{\boldsymbol h}_x^{\top}\boldsymbol W_k\tilde{\boldsymbol f}+ \boldsymbol b_k,
%   \]
%   where $\boldsymbol{W}_k \in \mathbb{R}^{d \times d}$ is the $k$-th slice and $\boldsymbol b_k$ is a bias term.
  
%   \noindent\textbf{MLP.}\;  
%   A lightweight two-layer perceptron on the concatenation $[\tilde{\boldsymbol h}_x\parallel \tilde{\boldsymbol f}]$:
%   \[
%   \widehat{\boldsymbol y}_x
%         =\sigma \!\left(\boldsymbol W_2 \cdot \sigma\!\left(\boldsymbol W_1[\tilde{\boldsymbol h}_x\parallel \tilde{\boldsymbol f}]
%                          +\boldsymbol b_1\right)+\boldsymbol b_2\right).
%   \]

\noindent\textbf{Edge predictor.}\;
Finally, any differentiable decoder $g(\cdot,\cdot)$ (e.g., dot-product or MLP) converts the fused embeddings into a link probability:
\[
\widehat{p}_{uv}(t)=\pi \bigl(g(\widehat{\boldsymbol y}_u,\widehat{\boldsymbol y}_v)\bigr),
\]
where $\pi$ is the sigmoid.  
Because only the input embeddings are augmented with vision cues, the downstream link prediction layer remains unchanged.

\section{Experiments}
\label{lab: experiment}
\begin{table*}[!ht]
\centering
\small
% \footnotesize
\setlength{\tabcolsep}{1mm}{
\begin{tabular}{l|ccccccccc}
\toprule
NSS & Datasets & TGN & CAWN & TCL & GraphMixer & DyGFormer & DyGFormer+ & TPNet & TPNet+ \\
\midrule
 & Reddit & $97.50_{ \pm 0.07}$ & $98.62_{ \pm 0.01}$ & $94.09_{ \pm 0.07}$ & $95.26_{ \pm 0.02}$ & $98.84_{ \pm 0.02}$ & \underline{$98.86_{ \pm 0.02}$}  & \underline{$98.86_{ \pm 0.01}$} & \textbf{98.95}$_{ \pm 0.03}$ \\
 & MOOC & $89.04_{ \pm 1.17}$ & $81.42_{ \pm 0.24}$ & $82.38_{ \pm 0.24}$ & $81.41_{ \pm 0.21}$ & $86.96_{ \pm 0.43}$ & $86.92_{ \pm 0.28}$ & \underline{$95.07_{ \pm 0.26}$} & \textbf{95.10}$_{ \pm 0.10}$ \\
rnd & Enron & $77.94_{ \pm 1.02}$ & $86.35_{ \pm 0.51}$ & $79.70_{ \pm 0.71}$ & $75.88_{ \pm 0.48}$ & $89.76_{ \pm 0.34}$ & $89.88_{ \pm 0.12}$ & \textbf{90.34}$_{ \pm 0.28}$ & \underline{$90.18_{ \pm 0.27}$} \\
 & UCI & $88.12_{ \pm 2.05}$ & $92.73_{ \pm 0.06}$ & $89.57_{ \pm 1.63}$ & $91.19_{ \pm 0.42}$ & $94.54_{ \pm 0.12}$ & $94.89_{ \pm 0.04}$ & \underline{$95.74_{ \pm 0.05}$} & \textbf{95.83}$_{ \pm 0.16}$ \\
 & Can. Parl. & $70.88_{ \pm 2.34}$ & $69.82_{ \pm 2.34}$ & $68.67_{ \pm 2.67}$ & $77.04_{ \pm 0.46}$ & $87.74_{ \pm 0.71}$ & \underline{$87.89_{ \pm 1.01}$} & $68.09_{ \pm 1.55}$ & \textbf{95.50}$_{ \pm 0.49}$ \\
 \cmidrule{2-10}
& Avg. rank & 5.80 & 5.60 & 6.80 & 6.60 & 3.80 & 3.00 & 3.00 & 1.20 \\
\midrule
 & Reddit & $64.85_{ \pm 0.85}$ & $63.67_{ \pm 0.41}$ & $60.83_{ \pm 0.25}$ & $64.50_{ \pm 0.26}$ & \textbf{65.37}$_{ \pm 0.60}$ & \underline{$65.07_{ \pm 0.91}$} & $62.15_{ \pm 1.72}$ & $64.88_{ \pm 2.21}$ \\
 & MOOC & $77.07_{ \pm 3.41}$ & $74.68_{ \pm 0.68}$ & $77.06_{ \pm 0.41}$ & $74.00_{ \pm 0.97}$ & $80.82_{ \pm 0.30}$ & \underline{$82.17_{ \pm 1.31}$} & $81.85_{ \pm 1.60}$ & \textbf{83.96}$_{ \pm 0.91}$ \\
hist & Enron & $62.91_{ \pm 1.16}$ & $60.70_{ \pm 0.36}$ & $70.66_{ \pm 0.39}$ &$72.37_{ \pm 1.37}$ & $67.07_{ \pm 0.62}$ & $67.20_{ \pm 1.19}$ & \underline{$74.60_{ \pm 1.35}$} & \textbf{76.19}$_{ \pm 1.14}$ \\
 & UCI & $70.78_{ \pm 0.78}$ & $64.54_{ \pm 0.47}$ & \underline{$80.25_{ \pm 0.39}$} & \textbf{81.66}$_{ \pm 0.49}$ & $72.13_{ \pm 1.87}$ & $72.23_{ \pm 1.49}$ & $78.48_{ \pm 1.18}$ & 79.57$_{ \pm 1.33}$ \\
 & Can. Parl. & $68.42_{ \pm 3.07}$ & $66.53_{ \pm 2.77}$ & $65.93_{ \pm 3.00}$ & $74.34_{ \pm 0.87}$ & $87.40_{ \pm 0.85}$ & \underline{$87.54_{ \pm 0.57}$} & $68.97_{ \pm 1.60}$ & \textbf{95.20}$_{ \pm 0.57}$ \\
 \cmidrule{2-10}
  & Avg. rank & 5.80 & 7.20 & 5.60 & 4.20 & 4.00 & 3.20 & 4.20 & 1.80 \\
\midrule
 & Reddit & $64.84_{ \pm 0.84}$ & $63.65_{ \pm 0.41}$ & $60.81_{ \pm 0.26}$ & $64.49_{ \pm 0.25}$ & \textbf{65.35}$_{ \pm 0.60}$ & $64.66_{ \pm 0.12}$ & $62.14_{ \pm 1.72}$ & \underline{$64.88_{ \pm 2.22}$} \\
 & MOOC & $77.07_{ \pm 3.40}$ & $74.69_{ \pm 0.68}$ & $74.65_{ \pm 0.54}$ & $73.99_{ \pm 0.97}$ & $80.82_{ \pm 0.30}$ & \underline{$82.17_{ \pm 1.30}$} & $81.85_{ \pm 1.60}$ & \textbf{83.96}$_{ \pm 0.91}$ \\
ind & Enron & $62.90_{ \pm 1.16}$ & $60.72_{ \pm 0.36}$ & $71.29_{ \pm 0.32}$ & $72.37_{ \pm 1.38}$ & $67.07_{ \pm 0.62}$ & $67.20_{ \pm 1.19}$ & \underline{$74.60_{ \pm 1.35}$} & \textbf{76.19}$_{ \pm 1.14}$ \\
 & UCI & $70.73_{ \pm 0.79}$ & $64.54_{ \pm 0.47}$ & $76.01_{ \pm 1.11}$ & \textbf{81.64}$_{ \pm 0.49}$ & $72.13_{ \pm 1.86}$ & $72.28_{ \pm 1.49}$ & $78.50_{ \pm 1.18}$ & \underline{$79.52_{ \pm 1.56}$} \\
 & Can. Parl. & $65.34_{ \pm 2.87}$ & $67.75_{ \pm 1.00}$ & $65.85_{ \pm 1.75}$ & $69.48_{ \pm 0.63}$ & $87.22_{ \pm 0.82}$ & \underline{$87.78_{ \pm 0.67}$} & $68.87_{ \pm 1.68}$ & \textbf{95.25}$_{ \pm 0.56}$\\
 \cmidrule{2-10}
  & Avg. rank & 6.00 & 6.80 & 6.00 & 4.20 & 4.00 & 3.60 & 4.00 & 1.40 \\
\bottomrule
\end{tabular}}
\caption{AP (\%) for \textit{inductive} dynamic link prediction on real-world datasets with three sampling strategies (NSS).}
% \td{to avoid second line with just (NSS) can swap ``real-world'' with ``five'' or even remove the comment on datasets. else, if wanting to keep the full amount here (ie not reduce to a single clean line), can shift the comment of bold and underline from the main text to fig caption }
\label{tab2}
\end{table*}
\subsection{Experimental Settings}
\paragraph{Datasets and Baselines.} We conduct experiments on five real-world datasets: Reddit, MOOC, Enron, Can. Parl., and UCI, all collected by~\cite{poursafaei2022towards}, and covering diverse domains. To validate the effectiveness of our model, we compare against six popular dynamic graph learning baselines based on various techniques, including memory networks (i.e., TGN~\cite{rossi2020temporal}), random walks (i.e., CAWN~\cite{wang2021inductive} and TPNet~\cite{lu2024improving}), MLP models (i.e., Graph-Mixer~\cite{cong2023we}), and sequential models (i.e., TCL~\cite{wang2021tcl} and DyGFormer~\cite{yu2023towards}). In addition, we integrate our approach into DyGFormer and TPNet to demonstrate the effectiveness and plug-and-play adaptability of Graph2Video. We refer to them as DyGFormer+ and TPNet+. Details about datasets and baselines can be found in Appendix A.

\paragraph{Evaluation Tasks and Metrics.} The evaluation of our experiment centers on the dynamic prediction task adopted in prior works~\cite{wu2019graph,rossi2020temporal,wang2021inductive,poursafaei2022towards}.
% This task predicts the probability of a link occurring between two given nodes at a specific time.
We utilize two settings: a transductive setting, where the objective is to predict future links between nodes observed during training, and an inductive setting, aiming to predict future links involving previously unseen nodes. Average Precision (AP) and Area Under the Receiver Operating Characteristic Curve (AUC-ROC) are chosen as the evaluation metrics. Similar to~\cite{poursafaei2022towards}, three negative sampling strategies (NSS) are used to evaluate our model, i.e., random (rnd), historical (hist), and inductive (ind). The latter two strategies are more challenging due to their inherent complexities, see~\cite{poursafaei2022towards} for details. We perform a chronological split of the dataset, assigning 70\% of the data to training, 15\% to validation, and 15\% to testing.

\paragraph{Implementation Details.} 
To ensure fair and consistent comparisons, we adopt the settings from~\cite{poursafaei2022towards}; details in Appendix B.
%To ensure fair and consistent performance comparisons, we follow the experimental settings and evaluation metrics reported in~\cite{poursafaei2022towards}, detailed in Appendix B. 
% \td{to save space. can just say, to ensure fair we follow X, details in appendix (then copy the rest of this paragraph to appendix)}

\subsection{Performance Comparison and Discussion}
Tables~\ref{tab1} and~\ref{tab2} present the AP scores of DyGFormer+ and TPNet+, compared against prior state-of-the-art dynamic graph models across five real-world datasets. All results are reported under transductive and inductive settings, evaluated with three negative-sampling strategies (random, historical, and inductive). 
For readability, all numerical values are multiplied by 100, with the highest and second-highest results in each column highlighted in \textbf{bold} and \underline{underline}, respectively. 
Additional results on AUC for both transductive and inductive link prediction tasks are provided in Appendix C.

\label{sec:ablation}
\begin{figure}[t]
% \small
\centering
\includegraphics[width=0.45\textwidth]{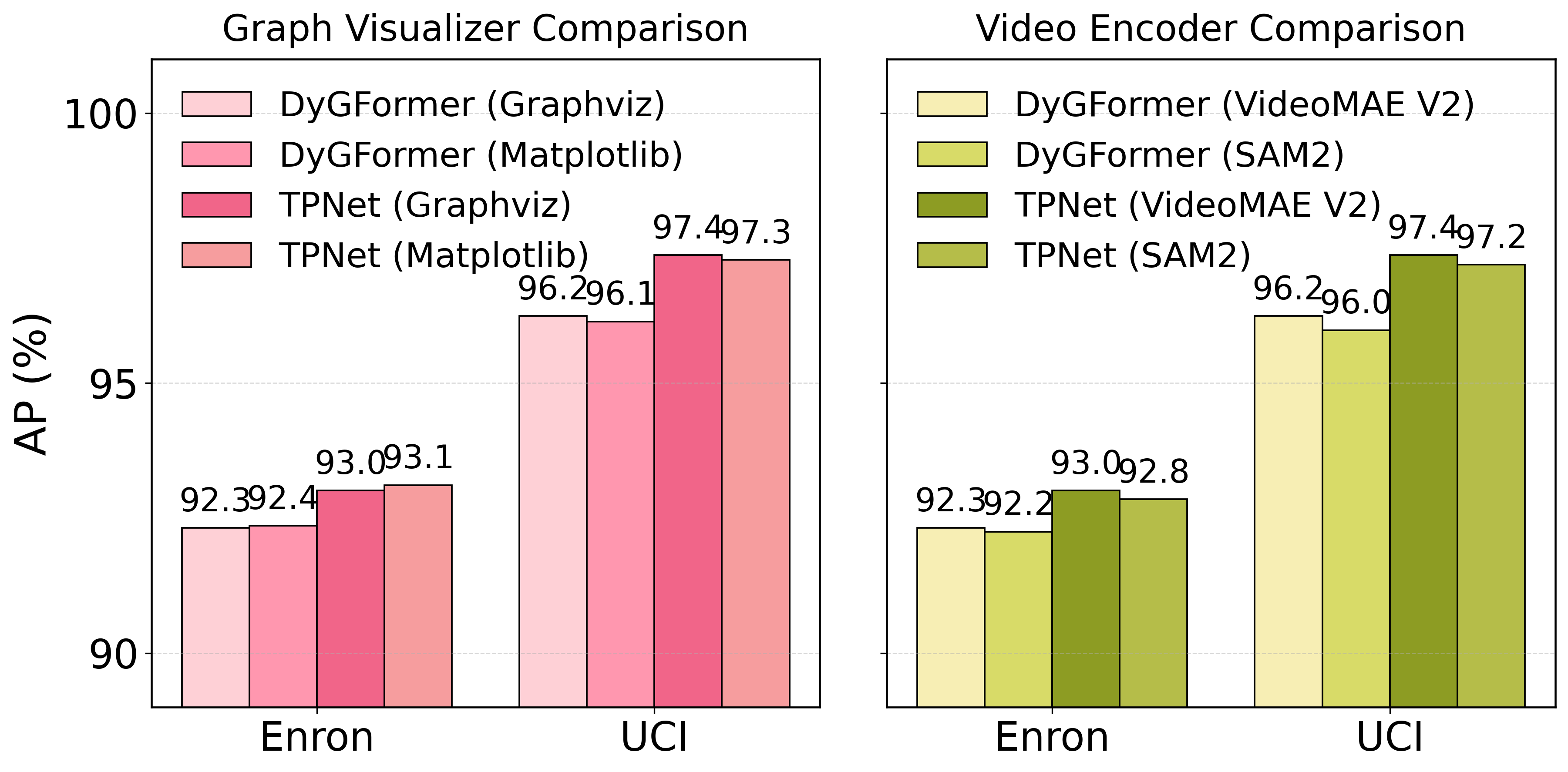} % Reduce the figure size so that it is slightly narrower than the column.
\caption{Graph Visualizer and Video Encoder Comparison.}
\label{fig: ap}
\end{figure}

\begin{table}[t]
\centering
\small
\begin{tabular}{c|c|cc}
\toprule
Model & s& Enron & UCI \\
\midrule
DyGFormer
& 8  & \textbf{92.35}$_{\pm 0.20}$ & 96.19$_{\pm 0.14}$ \\
& 16 & 92.32$_{\pm 0.18}$          & \textbf{96.24}$_{\pm 0.04}$ \\
& 32 & 92.31$_{\pm 0.16}$          & 96.10$_{\pm 0.38}$ \\
\midrule
TPNet
& 8  & \textbf{93.05}$_{\pm 0.18}$ & 97.23$_{\pm 0.21}$ \\
& 16 & 93.01$_{\pm 0.10}$          & \textbf{97.37}$_{\pm 0.10}$ \\
& 32 & 93.02$_{\pm 0.19}$          & 97.30$_{\pm 0.10}$ \\
\bottomrule
\end{tabular}
%\caption{Video frame with $k$ most recent neighbors performance comparison.}
\caption{Performance comparison when varying the number of most recent neighbors $s$ in the video frame.}
% \td{do we have results on smaller k here?}
\label{tab:recent neighbors}
\end{table}
% \paragraph{Comparison and Discussions.}
% Tables~\ref{tab1} and Table~\ref{tab2} present the AP scores of Graph2Video integrated with two dynamic graph backbones (TPNet and DyGFormer), compared against prior state-of-the-art CTDG models across five real-world datasets. 

The results demonstrate that integrating Graph2Video consistently enhances the performance of both backbones across nearly all datasets, sampling strategies, and transductive/inductive settings, thereby achieving new state-of-the-art
results on these benchmarks. The improvements are particularly evident under the inductive setting and challenging sampling strategies. For instance, TPNet+ achieves an average AP gain of \textbf{5.49\%} in the inductive setting with random sampling strategy across five datasets. Notably, under the historical and inductive sampling strategies for inductive link prediction, TPNet+ delivers average improvements of \textbf{6.75\%} and \textbf{6.77\%}, respectively, which modeling long‑range dependencies is especially challenging. The largest improvement is observed on the Can. Parl., with average AP gains of \textbf{10.88\%} in the transductive setting and \textbf{26.67\%} in the inductive setting across three sampling strategies. We attribute this remarkable boost primarily to the coarse‑grained annual partitioning scheme of the dataset, which diminishes fine‑grained temporal signals and increases reliance on long‑range dependencies and higher‑order neighborhood overlaps. We also report the average ranks (Avg. rank) of the models. The results show that integrated models consistently outperform the baselines. For example, TPNet+ achieves an average AP rank of 1.47 in both transductive and inductive settings across three sampling strategies. Thus, we can conclude that by capturing structural motions across aggregated frames, Graph2Video effectively restores local variations and global trends, thereby overcoming the limitations of current approaches.

\subsection{Ablation Study and Sensitivity Analysis}
\paragraph{Graph Visualizer Selection.}
Figure~\ref{fig: ap} (left) shows the performance
with different graph visualizers. The results demonstrate
stable performance across %various 
graph visualizers.

\begin{table}[t]
\centering
\small
\setlength{\tabcolsep}{1mm}{
\begin{tabular}{c|c|cc}
\toprule
Model & Fusion Strategy &Enron & UCI \\
\midrule
DyGFormer
& MLP       & \textbf{92.40}$_{\pm 0.12}$ & 96.23$_{\pm 0.06}$ \\
& Bilinear  & 91.60$_{\pm 0.29}$          & 96.15$_{\pm 0.10}$ \\
& Attention & 92.32$_{\pm 0.18}$          & \textbf{96.24}$_{\pm 0.04}$ \\
\midrule
TPNet
& MLP       & 92.98$_{\pm 0.08}$          & 97.14$_{\pm 0.24}$ \\
& Bilinear  & \textbf{93.18}$_{\pm 0.12}$          & 97.04$_{\pm 0.21}$ \\
& Attention & 93.01$_{\pm 0.10}$ & \textbf{97.37}$_{\pm 0.10}$ \\
\bottomrule
\end{tabular}}
\caption{ Fusion strategies performance comparison.}
\label{tab:fusion strategies}
\end{table}

\begin{table}[t]
\centering
\small
\setlength{\tabcolsep}{1mm}{
\begin{tabular}{c|c|cc}
\toprule
Model &Sam2-F& Enron & UCI \\
\midrule
DyGFormer
& 8  & 92.22$_{\pm 0.13}$ &95.93$_{\pm 0.06}$  \\
& 16 & 92.25$_{\pm 0.13}$    & 95.98$_{\pm 0.08}$ \\
& 32 & \textbf{92.31}$_{\pm 0.14}$  & \textbf{96.01}$_{\pm 0.09}$  \\
\midrule
TPNet
& 8  &92.82$_{\pm 0.09}$  & 97.16$_{\pm 0.17}$\\
& 16 & 92.85$_{\pm 0.23}$          & 97.19$_{\pm 0.29}$ \\
& 32 &    \textbf{93.01}$_{\pm 0.11}$        & \textbf{97.29}$_{\pm 0.12}$ \\
\bottomrule
\end{tabular}}
\caption{Performance comparison when varying the number of Frames $F$ in the Sam2 video model.}
\label{tab:Frame}
\end{table}

% \begin{table}[t]
% \centering
% \small
% \begin{tabular}{c|c|cc}
% \toprule
% Model & k-hop& Enron & UCI \\
% \midrule
% DyGFormer
% & 1 & 92.32$_{\pm 0.18}$          & 96.24$_{\pm 0.04}$ \\
% & 2 &    \textbf{92.39}$_{\pm 0.11}$       &  \\
% \midrule
% TPNet 
% & 1 & 93.01$_{\pm 0.10}$          & 97.37$_{\pm 0.10}$  \\
% & 2 & \textbf{93.06}$_{\pm 0.15}$      &  \\
% \bottomrule
% \end{tabular}
% \label{tab:k-hop}
% \end{table}

\paragraph{Video Encoder Selection.}
Figure~\ref{fig: ap} (right) reports the performance with different video encoders. The results indicate that VideoMAE V2 consistently achieves superior effectiveness, owing to its advanced masked tube reconstruction strategy and dual masking design.

\paragraph{$s$-Most Recent Neighbors.}
Table~\ref{tab:recent neighbors} reports the performance with varying $s$ most recent neighbors for graph video construction, showing stable results across $s$ values. %different values of $s$.

\paragraph{Feature Fusion Strategy.}
Table~\ref{tab:fusion strategies} presents the performance for different feature fusion strategies. The experimental results indicate, in most cases, default attention-guided fusion outperforms both bilinear fusion and MLP fusion.

\paragraph{Number of Frames $F$.}
Table~\ref{tab:Frame} shows that larger $F$ consistently yields better performance, confirming that finer temporal granularity benefits Graph2Video.

\section{Conclusion}
We introduced Graph2Video, a novel “graph-as-video” viewpoint for dynamic graph learning.
By serializing the evolving neighborhood of each candidate link into a sequence of structural frames, our framework constructs a compact “graph video” representation that captures both local fine‑grained variations and long‑range temporal dependencies. This representation is processed through a lightweight, plug‑and‑play module that derives link‑level spatio‑temporal embeddings, making the approach broadly applicable to diverse dynamic graph encoders without architectural modifications.
Comprehensive experiments on five benchmarks demonstrate that Graph2Video consistently improves strong baselines and achieves state-of-the-art performance in most settings, including diverse sampling strategies and inductive scenarios.
Beyond link prediction, our findings indicate a broader opportunity: leveraging the mature inductive biases of video models can advance other dynamic graph tasks such as temporal motif discovery, anomaly detection, and event forecasting.

\section*{Acknowledgments}

This work was supported by the National Key Research and Development Program of China under grant number 2022YFA1004102 and National Natural Science Foundation of China (NSFC) under grant number 62136005.

\bibliography{aaai2026}

@article{poursafaei2022towards,
  title={Towards better evaluation for dynamic link prediction},
  author={Poursafaei, Farimah and Huang, Shenyang and Pelrine, Kellin and Rabbany, Reihaneh},
  journal={Advances in Neural Information Processing Systems},
  volume={35},
  pages={32928--32941},
  year={2022}
}

@article{gita,
  title={Gita: Graph to visual and textual integration for vision-language graph reasoning},
  author={Wei, Yanbin and Fu, Shuai and Jiang, Weisen and Zhang, Zejian and Zeng, Zhixiong and Wu, Qi and Kwok, James and Zhang, Yu},
  journal={Advances in Neural Information Processing Systems},
  volume={37},
  pages={44--72},
  year={2024}
}

@article{wei2025open,
  title={Open Your Eyes: Vision Enhances Message Passing Neural Networks in Link Prediction},
  author={Wei, Yanbin and Wang, Xuehao and Zhuang, Zhan and Chen, Yang and Chen, Shuhao and Zhang, Yulong and Zhang, Yu and Kwok, James},
  journal={arXiv preprint arXiv:2505.08266},
  year={2025}
}

@article{liu2020evolution,
  title={The evolution of structural balance in time-varying signed networks},
  author={Liu, Hua and Qu, Cunquan and Niu, Yawei and Wang, Guanghui},
  journal={Future Generation Computer Systems},
  volume={102},
  pages={403--408},
  year={2020},
  publisher={Elsevier}
}

@inproceedings{trivedi2019dyrep,
  title={Dyrep: Learning representations over dynamic graphs},
  author={Trivedi, Rakshit and Farajtabar, Mehrdad and Biswal, Prasenjeet and Zha, Hongyuan},
  booktitle={International conference on learning representations},
  year={2019}
}

@article{rossi2020temporal,
  title={Temporal graph networks for deep learning on dynamic graphs},
  author={Rossi, Emanuele and Chamberlain, Ben and Frasca, Fabrizio and Eynard, Davide and Monti, Federico and Bronstein, Michael},
  journal={ICML
2020 Workshop on Graph Representation Learning},
  year={2020}
}

@article{wu2019graph,
  title={Graph wavenet for deep spatial-temporal graph modeling},
  author={Wu, Zonghan and Pan, Shirui and Long, Guodong and Jiang, Jing and Zhang, Chengqi},
  journal={arXiv preprint arXiv:1906.00121},
  year={2019}
}

@article{wang2021inductive,
  title={Inductive representation learning in temporal networks via causal anonymous walks},
  author={Wang, Yanbang and Chang, Yen-Yu and Liu, Yunyu and Leskovec, Jure and Li, Pan},
  journal={arXiv preprint arXiv:2101.05974},
  year={2021}
}

@inproceedings{liu2021elastic,
  title={Elastic graph neural networks},
  author={Liu, Xiaorui and Jin, Wei and Ma, Yao and Li, Yaxin and Liu, Hua and Wang, Yiqi and Yan, Ming and Tang, Jiliang},
  booktitle={International Conference on Machine Learning},
  pages={6837--6849},
  year={2021},
  organization={PMLR}
}

@article{wang2021tcl,
  title={Tcl: Transformer-based dynamic graph modelling via contrastive learning},
  author={Wang, Lu and Chang, Xiaofu and Li, Shuang and Chu, Yunfei and Li, Hui and Zhang, Wei and He, Xiaofeng and Song, Le and Zhou, Jingren and Yang, Hongxia},
  journal={arXiv preprint arXiv:2105.07944},
  year={2021}
}

@inproceedings{liu2023enhancing,
  title={Enhancing graph representations learning with decorrelated propagation},
  author={Liu, Hua and Han, Haoyu and Jin, Wei and Liu, Xiaorui and Liu, Hui},
  booktitle={Proceedings of the 29th ACM SIGKDD Conference on Knowledge Discovery and Data Mining},
  pages={1466--1476},
  year={2023}
}

@article{cong2023we,
  title={Do we really need complicated model architectures for temporal networks?},
  author={Cong, Weilin and Zhang, Si and Kang, Jian and Yuan, Baichuan and Wu, Hao and Zhou, Xin and Tong, Hanghang and Mahdavi, Mehrdad},
  journal={International Conference on Learning Representations},
  year={2023}
}

@article{yu2023towards,
  title={Towards better dynamic graph learning: New architecture and unified library},
  author={Yu, Le and Sun, Leilei and Du, Bowen and Lv, Weifeng},
  journal={Advances in Neural Information Processing Systems},
  volume={36},
  pages={67686--67700},
  year={2023}
}

@inproceedings{fan2019graph,
  title={Graph neural networks for social recommendation},
  author={Fan, Wenqi and Ma, Yao and Li, Qing and He, Yuan and Zhao, Eric and Tang, Jiliang and Yin, Dawei},
  booktitle={The world wide web conference},
  pages={417--426},
  year={2019}
}

@article{zhang2022dynamic,
  title={Dynamic graph neural networks for sequential recommendation},
  author={Zhang, Mengqi and Wu, Shu and Yu, Xueli and Liu, Qiang and Wang, Liang},
  journal={IEEE Transactions on Knowledge and Data Engineering},
  volume={35},
  number={5},
  pages={4741--4753},
  year={2022},
  publisher={IEEE}
}

@inproceedings{gao2022graph,
  title={Graph neural networks for recommender system},
  author={Gao, Chen and Wang, Xiang and He, Xiangnan and Li, Yong},
  booktitle={Proceedings of the fifteenth ACM international conference on web search and data mining},
  pages={1623--1625},
  year={2022}
}

@inproceedings{wang2020traffic,
  title={Traffic flow prediction via spatial temporal graph neural network},
  author={Wang, Xiaoyang and Ma, Yao and Wang, Yiqi and Jin, Wei and Wang, Xin and Tang, Jiliang and Jia, Caiyan and Yu, Jian},
  booktitle={Proceedings of the web conference 2020},
  pages={1082--1092},
  year={2020}
}

@article{shu2017fake,
  title={Fake news detection on social media: A data mining perspective},
  author={Shu, Kai and Sliva, Amy and Wang, Suhang and Tang, Jiliang and Liu, Huan},
  journal={ACM SIGKDD explorations newsletter},
  volume={19},
  number={1},
  pages={22--36},
  year={2017},
  publisher={ACM New York, NY, USA}
}

@article{alvarez2021evolutionary,
  title={Evolutionary dynamics of higher-order interactions in social networks},
  author={Alvarez-Rodriguez, Unai and Battiston, Federico and de Arruda, Guilherme Ferraz and Moreno, Yamir and Perc, Matja{\v{z}} and Latora, Vito},
  journal={Nature Human Behaviour},
  volume={5},
  number={5},
  pages={586--595},
  year={2021},
  publisher={Nature Publishing Group UK London}
}

@inproceedings{kumar2019predicting,
  title={Predicting dynamic embedding trajectory in temporal interaction networks},
  author={Kumar, Srijan and Zhang, Xikun and Leskovec, Jure},
  booktitle={Proceedings of the 25th ACM SIGKDD international conference on knowledge discovery \& data mining},
  pages={1269--1278},
  year={2019}
}

@article{bui2022spatial,
  title={Spatial-temporal graph neural network for traffic forecasting: An overview and open research issues},
  author={Bui, Khac-Hoai Nam and Cho, Jiho and Yi, Hongsuk},
  journal={Applied Intelligence},
  volume={52},
  number={3},
  pages={2763--2774},
  year={2022},
  publisher={Springer}
}

@article{sharma2023graph,
  title={A graph neural network (GNN)-based approach for real-time estimation of traffic speed in sustainable smart cities},
  author={Sharma, Amit and Sharma, Ashutosh and Nikashina, Polina and Gavrilenko, Vadim and Tselykh, Alexey and Bozhenyuk, Alexander and Masud, Mehedi and Meshref, Hossam},
  journal={Sustainability},
  volume={15},
  number={15},
  pages={11893},
  year={2023},
  publisher={MDPI}
}

@article{kipf2016semi,
  title={Semi-supervised classification with graph convolutional networks},
  author={Kipf, Thomas N and Welling, Max},
  journal={arXiv preprint arXiv:1609.02907},
  year={2016}
}

@article{hamilton2017inductive,
  title={Inductive representation learning on large graphs},
  author={Hamilton, Will and Ying, Zhitao and Leskovec, Jure},
  journal={Advances in neural information processing systems},
  volume={30},
  year={2017}
}

@article{velivckovic2017graph,
  title={Graph attention networks},
  author={Veli{\v{c}}kovi{\'c}, Petar and Cucurull, Guillem and Casanova, Arantxa and Romero, Adriana and Lio, Pietro and Bengio, Yoshua},
  journal={arXiv preprint arXiv:1710.10903},
  year={2017}
}

@article{cong2021dynamic,
  title={Dynamic graph representation learning via graph transformer networks},
  author={Cong, Weilin and Wu, Yanhong and Tian, Yuandong and Gu, Mengting and Xia, Yinglong and Mahdavi, Mehrdad and Chen, Chun-cheng Jason},
  journal={OpenReview},
  year={2022}
}

@inproceedings{sankar2020dysat,
  title={Dysat: Deep neural representation learning on dynamic graphs via self-attention networks},
  author={Sankar, Aravind and Wu, Yanhong and Gou, Liang and Zhang, Wei and Yang, Hao},
  booktitle={Proceedings of the 13th international conference on web search and data mining},
  pages={519--527},
  year={2020}
}

@inproceedings{you2022roland,
  title={ROLAND: graph learning framework for dynamic graphs},
  author={You, Jiaxuan and Du, Tianyu and Leskovec, Jure},
  booktitle={Proceedings of the 28th ACM SIGKDD conference on knowledge discovery and data mining},
  pages={2358--2366},
  year={2022}
}

@article{souza2022provably,
  title={Provably expressive temporal graph networks},
  author={Souza, Amauri and Mesquita, Diego and Kaski, Samuel and Garg, Vikas},
  journal={Advances in neural information processing systems},
  volume={35},
  pages={32257--32269},
  year={2022}
}

@article{su2024pres,
  title={Pres: Toward scalable memory-based dynamic graph neural networks},
  author={Su, Junwei and Zou, Difan and Wu, Chuan},
  journal={arXiv preprint arXiv:2402.04284},
  year={2024}
}

@inproceedings{sheng2024mspipe,
  title={Mspipe: Efficient temporal gnn training via staleness-aware pipeline},
  author={Sheng, Guangming and Su, Junwei and Huang, Chao and Wu, Chuan},
  booktitle={Proceedings of the 30th ACM SIGKDD Conference on Knowledge Discovery and Data Mining},
  pages={2651--2662},
  year={2024}
}

@inproceedings{ji2024memmap,
  title={Memmap: An adaptive and latent memory structure for dynamic graph learning},
  author={Ji, Shuo and Liu, Mingzhe and Sun, Leilei and Liu, Chuanren and Zhu, Tongyu},
  booktitle={Proceedings of the 30th ACM SIGKDD Conference on Knowledge Discovery and Data Mining},
  pages={1257--1268},
  year={2024}
}

@article{li2023zebra,
  title={Zebra: When temporal graph neural networks meet temporal personalized pagerank},
  author={Li, Yiming and Shen, Yanyan and Chen, Lei and Yuan, Mingxuan},
  journal={Proceedings of the VLDB Endowment},
  volume={16},
  number={6},
  pages={1332--1345},
  year={2023},
  publisher={VLDB Endowment}
}

@article{xu2020inductive,
  title={Inductive representation learning on temporal graphs},
  author={Xu, Da and Ruan, Chuanwei and Korpeoglu, Evren and Kumar, Sushant and Achan, Kannan},
  journal={arXiv preprint arXiv:2002.07962},
  year={2020}
}

@article{schuster1997bidirectional,
  title={Bidirectional recurrent neural networks},
  author={Schuster, Mike and Paliwal, Kuldip K},
  journal={IEEE transactions on Signal Processing},
  volume={45},
  number={11},
  pages={2673--2681},
  year={1997},
  publisher={Ieee}
}

@article{tong2022videomae,
  title={Videomae: Masked autoencoders are data-efficient learners for self-supervised video pre-training},
  author={Tong, Zhan and Song, Yibing and Wang, Jue and Wang, Limin},
  journal={Advances in neural information processing systems},
  volume={35},
  pages={10078--10093},
  year={2022}
}

@inproceedings{wang2023videomae,
  title={Videomae v2: Scaling video masked autoencoders with dual masking},
  author={Wang, Limin and Huang, Bingkun and Zhao, Zhiyu and Tong, Zhan and He, Yinan and Wang, Yi and Wang, Yali and Qiao, Yu},
  booktitle={Proceedings of the IEEE/CVF conference on computer vision and pattern recognition},
  pages={14549--14560},
  year={2023}
}

@article{ravi2024sam,
  title={Sam 2: Segment anything in images and videos},
  author={Ravi, Nikhila and Gabeur, Valentin and Hu, Yuan-Ting and Hu, Ronghang and Ryali, Chaitanya and Ma, Tengyu and Khedr, Haitham and R{\"a}dle, Roman and Rolland, Chloe and Gustafson, Laura and others},
  journal={arXiv preprint arXiv:2408.00714},
  year={2024}
}

@inproceedings{arnab2021vivit,
  title={Vivit: A video vision transformer},
  author={Arnab, Anurag and Dehghani, Mostafa and Heigold, Georg and Sun, Chen and Lu{\v{c}}i{\'c}, Mario and Schmid, Cordelia},
  booktitle={Proceedings of the IEEE/CVF international conference on computer vision},
  pages={6836--6846},
  year={2021}
}

@article{gansner2000open,
  title={An open graph visualization system and its applications to software engineering},
  author={Gansner, Emden R and North, Stephen C},
  journal={Software: practice and experience},
  volume={30},
  number={11},
  pages={1203--1233},
  year={2000},
  publisher={Wiley Online Library}
}

@book{tosi2009matplotlib,
  title={Matplotlib for Python developers},
  author={Tosi, Sandro},
  volume={307},
  year={2009},
  publisher={Packt Publishing Birmingham, UK}
}

@article{barros2021survey,
  title={A survey on embedding dynamic graphs},
  author={Barros, Claudio DT and Mendon{\c{c}}a, Matheus RF and Vieira, Alex B and Ziviani, Artur},
  journal={ACM Computing Surveys (CSUR)},
  volume={55},
  number={1},
  pages={1--37},
  year={2021},
  publisher={ACM New York, NY}
}

@article{kazemi2020representation,
  title={Representation learning for dynamic graphs: A survey},
  author={Kazemi, Seyed Mehran and Goel, Rishab and Jain, Kshitij and Kobyzev, Ivan and Sethi, Akshay and Forsyth, Peter and Poupart, Pascal},
  journal={Journal of Machine Learning Research},
  volume={21},
  number={70},
  pages={1--73},
  year={2020}
}

@article{peng2025tidformer,
  title={TIDFormer: Exploiting Temporal and Interactive Dynamics Makes A Great Dynamic Graph Transformer},
  author={Peng, Jie and Wei, Zhewei and Ye, Yuhang},
  journal={arXiv preprint arXiv:2506.00431},
  year={2025}
}

@article{vaswani2017attention,
  title={Attention is all you need},
  author={Vaswani, Ashish and Shazeer, Noam and Parmar, Niki and Uszkoreit, Jakob and Jones, Llion and Gomez, Aidan N and Kaiser, {\L}ukasz and Polosukhin, Illia},
  journal={Advances in neural information processing systems},
  volume={30},
  year={2017}
}

@inproceedings{gberta_2021_ICML,
    author  = {Gedas Bertasius and Heng Wang and Lorenzo Torresani},
    title = {Is Space-Time Attention All You Need for Video Understanding?},
    booktitle   = {Proceedings of the International Conference on Machine Learning (ICML)}, 
    month = {July},
    year = {2021}
}

@inproceedings{fan2021multiscale,
  title={Multiscale vision transformers},
  author={Fan, Haoqi and Xiong, Bo and Mangalam, Karttikeya and Li, Yanghao and Yan, Zhicheng and Malik, Jitendra and Feichtenhofer, Christoph},
  booktitle={Proceedings of the IEEE/CVF international conference on computer vision},
  pages={6824--6835},
  year={2021}
}

@article{lu2024improving,
  title={Improving temporal link prediction via temporal walk matrix projection},
  author={Lu, Xiaodong and Sun, Leilei and Zhu, Tongyu and Lv, Weifeng},
  journal={Advances in Neural Information Processing Systems},
  volume={37},
  pages={141153--141182},
  year={2024}
}

@article{ma2024mixture,
  title={Mixture of link predictors on graphs},
  author={Ma, Li and Han, Haoyu and Li, Juanhui and Shomer, Harry and Liu, Hui and Gao, Xiaofeng and Tang, Jiliang},
  journal={Advances in Neural Information Processing Systems},
  volume={37},
  pages={16043--16070},
  year={2024}
}

@inproceedings{wang2024internvideo2,
  title={Internvideo2: Scaling foundation models for multimodal video understanding},
  author={Wang, Yi and Li, Kunchang and Li, Xinhao and Yu, Jiashuo and He, Yinan and Chen, Guo and Pei, Baoqi and Zheng, Rongkun and Wang, Zun and Shi, Yansong and others},
  booktitle={European Conference on Computer Vision},
  pages={396--416},
  year={2024},
  organization={Springer}
}

% %%%%%%%%%%%%%%%%%%%%%%%%%%%%%%%%%%%%%%%%%%%%%%%%%%%%%%%%%%%%

% % Check whether the conference requires a reproducibility checklist to be included in the paper.
% % If so, you can uncomment the following line and ajust the path to include it.
% \input{../../ReproducibilityChecklist/LaTeX/ReproducibilityChecklist.tex}
\clearpage
\appendix

% \clearpage
% \begin{center}
%     \Large \textbf{Technical Appendix for “Graph2Video: Leveraging Video Models to Model Dynamic Graph Evolution”}
% \end{center}
\begin{table*}[t]
\centering
\begin{tabular}{l l r r c c l r l}
\toprule
Datasets       & Domains        & \#Nodes   & \#Links     & \#N\&L Feat
                          & Bipartite & Duration   & Unique Steps & Time Granularity \\
\midrule
Reddit        & Social      & 10,984  & 672,447   & – \& 172      & True        & 1 month     & 669,065     & Unix timestamps \\
MOOC          & Interaction & 7,144   & 411,749   & – \& 4        & True        & 17 month     & 345,600     & Unix timestamps \\
Enron         & Social      & 184     & 125,235   & – \& –        & False         & 3 years   & 22,632      & Unix timestamps \\
UCI           & Social      & 1,899   & 59,835    & – \& –        & False         & 196 days    & 58,911      & Unix timestamps \\
Can. Parl.   & Politics   & 734     & 74,478 & – \& 1        & False        &  14 years    & 14     &  years \\
\bottomrule
\end{tabular}
\caption{Statistics of the five real-world datasets.}
\label{tbl:dataset-stats}
\end{table*}

\begin{center}
    \Large \textbf{Appendix}
\end{center}
\renewcommand{\thetable}{A\arabic{table}}
\setcounter{table}{0} 
\renewcommand{\thefigure}{A\arabic{figure}}
\setcounter{figure}{0}

\section{Datasets and Baselines}
\setcounter{table}{6} % 如果正文最后一个表是 Table 6
\renewcommand{\thetable}{\arabic{table}}
\begin{table*}[t]
\centering
\resizebox{\textwidth}{!}{
\begin{tabular}{llcccccccc}
\toprule
NSS & Datasets & TGN & CAWN & TCL & GraphMixer & DyGFormer & DyGFormer+ & TPNet & TPNet+ \\
\midrule
 & Reddit & $98.60_{\pm 0.06}$ & $99.01_{\pm 0.01}$ & $97.42_{\pm 0.02}$ & $97.17_{\pm 0.02}$ & $99.15_{\pm 0.01}$ & $99.16_{\pm 0.01}$ & \underline{$99.22_{\pm 0.00}$} & $\textbf{99.27}_{\pm 0.01}$ \\
 & MOOC & $91.21_{\pm 1.15}$ & $80.38_{\pm 0.26}$ & $83.12_{\pm 0.18}$ & $84.01_{\pm 0.17}$ & $87.91_{\pm 0.58}$ & $86.93_{\pm 0.38}$ & $\textbf{97.17}_{\pm 0.08}$ & \underline{$97.13_{\pm 0.02}$} \\
rnd & Enron & $88.32_{\pm 0.99}$ & $90.45_{\pm 0.14}$ & $75.74_{\pm 0.72}$ & $84.38_{\pm 0.21}$ & $93.33_{\pm 0.13}$ & $93.45_{\pm 0.13}$ & \underline{$93.98_{\pm 0.26}$} & $\textbf{94.16}_{\pm 0.10}$ \\
 & UCI & $92.03_{\pm 1.13}$ & $93.87_{\pm 0.08}$ & $87.82_{\pm 1.36}$ & $91.81_{\pm 0.67}$ & $94.49_{\pm 0.26}$ & $95.22_{\pm 0.07}$ & \underline{$96.79_{\pm 0.05}$} & $\textbf{96.86}_{\pm 0.14}$ \\
 & Can. Parl. & $76.99_{\pm 1.80}$ & $75.70_{\pm 3.27}$ & $72.46_{\pm 3.23}$ & $83.17_{\pm 0.53}$ & $97.76_{\pm 0.41}$ & \underline{$97.99_{\pm 0.41}$} & $92.05_{\pm 0.34}$ & $\textbf{98.83}_{\pm 0.12}$ \\
 \midrule
 & Avg. rank & 5.40 & 6.00 & 7.60 & 6.60 & 3.80 & 3.20 & 2.20 & 1.20 \\
\midrule
 & Reddit & $81.11_{\pm 0.19}$ & $80.27_{\pm 0.30}$ & $76.49_{\pm 0.16}$ & $77.80_{\pm 0.12}$ & $80.54_{\pm 0.29}$ & $80.63_{\pm 0.88}$ & $\textbf{81.87}_{\pm 0.49}$ & \underline{$81.79_{\pm 0.50}$} \\
 & MOOC & $88.00_{\pm 1.80}$ & $71.57_{\pm 1.07}$ & $72.09_{\pm 0.56}$ & $76.68_{\pm 1.40}$ & $87.04_{\pm 0.35}$ & $88.13_{\pm 0.62}$ & \underline{$93.45_{\pm 0.67}$} & $\textbf{93.60}_{\pm 0.25}$ \\
hist & Enron & $77.09_{\pm 2.22}$ & $65.10_{\pm 0.34}$ & $67.95_{\pm 0.88}$ & $75.27_{\pm 1.14}$ & $76.55_{\pm 0.52}$ & $77.76_{\pm 0.43}$ & \underline{$81.16_{\pm 1.28}$} & $\textbf{82.14}_{\pm 0.96}$ \\
 & UCI & $77.25_{\pm 2.68}$ & $57.86_{\pm 0.15}$ & $72.25_{\pm 3.46}$ & $77.54_{\pm 2.02}$ & $76.97_{\pm 0.24}$ & $78.19_{\pm 0.60}$ & \underline{$80.42_{\pm 0.64}$} & $\textbf{80.50}_{\pm 0.80}$ \\
 & Can. Parl. & $73.23_{\pm 3.08}$ & $72.06_{\pm 3.94}$ & $69.95_{\pm 3.70}$ & $79.03_{\pm 1.01}$ & $97.61_{\pm 0.40}$ & $\textbf{98.02}_{\pm 0.40}$ & $86.39_{\pm 3.73}$ & \underline{$97.64_{\pm 0.32}$} \\
 \midrule
 & Avg. rank & 4.40 & 7.40 & 7.40 & 5.60 & 4.80 & 2.80 & 2.20 & 1.40 \\
\midrule
 & Reddit & $84.56_{\pm 0.29}$ & $\textbf{88.04}_{\pm 0.29}$ & $84.67_{\pm 0.29}$ & $82.21_{\pm 0.13}$ & $86.23_{\pm 0.51}$ & \underline{$86.55_{\pm 0.63}$} & $81.64_{\pm 0.42}$ & \textbf{$82.49_{\pm 1.51}$} \\
 & MOOC & $77.44_{\pm 2.86}$ & $70.32_{\pm 1.43}$ & $70.36_{\pm 0.37}$ & $72.45_{\pm 0.72}$ & $80.76_{\pm 0.76}$ & $81.10_{\pm 1.42}$ & \underline{$89.07_{\pm 0.63}$} & $\textbf{89.23}_{\pm 0.02}$ \\
ind & Enron & $71.34_{\pm 2.46}$ & $75.17_{\pm 0.50}$ & $67.64_{\pm 0.86}$ & $71.53_{\pm 0.85}$ & $74.07_{\pm 0.64}$ & $75.08_{\pm 0.49}$ & \underline{$75.44_{\pm 1.38}$} & $\textbf{76.40}_{\pm 0.96}$ \\
 & UCI & $64.11_{\pm 1.04}$ & $58.06_{\pm 0.26}$ & $70.05_{\pm 1.86}$ & $\textbf{74.59}_{\pm 0.74}$ & $65.96_{\pm 1.18}$ & $66.09_{\pm 1.06}$ & $70.85_{\pm 0.96}$ & \underline{$71.62_{\pm 1.02}$}\\
 & Can. Parl. & $69.57_{\pm 2.81}$ & $72.93_{\pm 1.78}$ & $69.47_{\pm 2.12}$ & $70.52_{\pm 0.94}$ & $96.70_{\pm 0.59}$ & \underline{$97.02_{\pm 0.39}$} & $85.05_{\pm 2.71}$ & $\textbf{98.82}_{\pm 0.13}$ \\
 \midrule
 & Avg. rank & 6.20 & 5.00 & 6.20 & 5.20 & 4.20 & 3.20 & 3.80 & 2.20 \\
\bottomrule
\end{tabular}}
\caption{AUC (\%) for transductive dynamic link prediction on real-world datasets with three sampling strategies (NSS).}
\label{tab: AUC_trans}
\end{table*}

\begin{table*}[!t]
\centering
\resizebox{\textwidth}{!}{
\begin{tabular}{llcccccccc}
\toprule
 NSS & Datasets & TGN & CAWN & TCL & GraphMixer & DyGFormer & DyGFormer+ & TPNet & TPNet+ \\
\midrule
 & Reddit & $97.39_{\pm 0.07}$ & $98.42_{\pm 0.02}$ & $93.80_{\pm 0.07}$ & $94.97_{\pm 0.05}$ & $98.71_{\pm 0.01}$ & $98.71_{\pm 0.01}$ & $\underline{98.73_{\pm 0.02}}$ & $\textbf{98.83}_{\pm 0.03}$ \\
 & MOOC   & $91.24_{\pm 0.99}$ & $81.86_{\pm 0.25}$ & $81.43_{\pm 0.19}$ & $82.77_{\pm 0.24}$ & $87.62_{\pm 0.51}$ & $87.91_{\pm 0.24}$ & $\textbf{95.55}_{\pm 0.25}$ & $\textbf{95.55}_{\pm 0.15}$ \\
rnd & Enron & $78.83_{\pm 1.11}$ & $87.02_{\pm 0.50}$ & $72.33_{\pm 0.99}$ & $76.51_{\pm 0.71}$ & $\underline{90.69_{\pm 0.26}}$ & $\textbf{90.78}_{\pm 0.17}$ & $90.21_{\pm 0.49}$ & $90.18_{\pm 0.48}$ \\
 & UCI    & $86.68_{\pm 2.29}$ & $90.40_{\pm 0.11}$ & $84.49_{\pm 1.82}$ & $89.30_{\pm 0.57}$ & $92.63_{\pm 0.13}$ & $93.27_{\pm 0.03}$ & $\underline{94.40_{\pm 0.03}}$ & $\textbf{94.52}_{\pm 0.26}$ \\
 & Can. Parl. & $55.86_{\pm 0.75}$ & $58.83_{\pm 1.13}$ & $55.83_{\pm 1.07}$ & $58.32_{\pm 1.08}$ & $89.33_{\pm 0.48}$ & $\underline{89.51_{\pm 0.82}}$ & $69.21_{\pm 1.31}$ & $\textbf{95.69}_{\pm 0.50}$ \\
\midrule
 & Avg. rank & 5.80 & 5.40 & 8.00 & 6.40 & 3.40 & 2.80 & 2.60 & 1.60 \\
\midrule
 & Reddit & $64.55_{\pm 0.50}$ & $\textbf{64.94}_{\pm 0.21}$ & $61.43_{\pm 0.26}$ & $64.27_{\pm 0.13}$ & $64.81_{\pm 0.25}$ & $\underline{64.82_{\pm 0.52}}$ & $62.37_{\pm 0.83}$ & $63.58_{\pm 0.83}$ \\
 & MOOC   & $77.69_{\pm 3.55}$ & $71.68_{\pm 0.94}$ & $69.82_{\pm 0.32}$ & $72.53_{\pm 0.84}$ & $80.77_{\pm 0.63}$ & $81.32_{\pm 1.05}$ & $\underline{84.46_{\pm 0.88}}$ & $\textbf{85.12}_{\pm 0.46}$ \\
hist & Enron & $62.68_{\pm 1.09}$ & $62.25_{\pm 0.40}$ & $64.06_{\pm 1.02}$ & $68.20_{\pm 1.62}$ & $65.78_{\pm 0.42}$ & $66.27_{\pm 0.35}$ & $\underline{74.50_{\pm 1.02}}$ & $\textbf{75.13}_{\pm 0.93}$ \\
 & UCI    & $62.69_{\pm 0.90}$ & $56.39_{\pm 0.10}$ & $70.46_{\pm 1.94}$ & $\textbf{75.98}_{\pm 0.84}$ & $65.55_{\pm 1.01}$ & $65.58_{\pm 0.94}$ & $71.35_{\pm 0.84}$ & $\underline{71.98_{\pm 1.10}}$ \\
 & Can. Parl. & $55.64_{\pm 0.54}$ & $60.11_{\pm 0.48}$ & $57.30_{\pm 1.03}$ & $56.68_{\pm 1.20}$ & $88.68_{\pm 0.74}$ & $\underline{88.91_{\pm 0.22}}$ & $69.11_{\pm 1.18}$ & $\textbf{95.41}_{\pm 0.58}$ \\
\midrule
& Avg. rank & 6.20 & 5.80 & 6.60 & 4.20 & 4.20 & 3.20 & 3.60 & 2.20 \\
\midrule
 & Reddit & $64.55_{\pm 0.50}$ & $\textbf{64.91}_{\pm 0.21}$ & $61.36_{\pm 0.26}$ & $64.27_{\pm 0.13}$ & $64.80_{\pm 0.25}$ & $\underline{64.80_{\pm 0.64}}$ & $62.37_{\pm 0.83}$ & $63.58_{\pm 0.84}$ \\
 & MOOC   & $77.68_{\pm 3.55}$ & $71.69_{\pm 0.94}$ & $69.83_{\pm 0.32}$ & $72.52_{\pm 0.84}$ & $80.77_{\pm 0.63}$ & $81.32_{\pm 1.05}$ & $\underline{84.46_{\pm 0.88}}$ & $\textbf{85.12}_{\pm 0.46}$ \\
ind & Enron & $62.68_{\pm 1.09}$ & $62.27_{\pm 0.40}$ & $64.05_{\pm 1.02}$ & $68.19_{\pm 1.63}$ & $65.79_{\pm 0.42}$ & $66.27_{\pm 0.35}$ & $\underline{74.50_{\pm 1.02}}$ & $\textbf{75.13}_{\pm 0.93}$ \\
 & UCI    & $62.66_{\pm 0.91}$ & $56.39_{\pm 0.11}$ & $70.42_{\pm 1.93}$ & $\textbf{75.97}_{\pm 0.85}$ & $65.58_{\pm 1.00}$ & $65.64_{\pm 0.95}$ & $71.37_{\pm 0.84}$ & $\underline{72.05_{\pm 1.36}}$ \\
 & Can. Parl. & $55.43_{\pm 0.42}$ & $60.01_{\pm 0.47}$ & $56.88_{\pm 0.93}$ & $56.63_{\pm 1.09}$ & $88.51_{\pm 0.73}$ & $\underline{89.12_{\pm 0.25}}$ & $68.98_{\pm 1.21}$ & $\textbf{95.46}_{\pm 0.57}$ \\
 \midrule
 & Avg. rank & 6.20 & 5.00 & 6.60 & 5.00 & 4.00 & 3.40 & 3.60 & 2.20 \\

\bottomrule
\end{tabular}}
\caption{AUC (\%) for inductive dynamic link prediction on real-world datasets with three sampling strategies (NSS).}
\label{tab: AUC_ind}
\end{table*}

\subsection{Datasets.}
We conduct experiments on five real-world datasets collected by~\cite{poursafaei2022towards}. The brief descriptions of these datasets are as follows:
\begin{itemize}[leftmargin=1.6em]
    \item \textbf{Reddit} is a bipartite graph that records user posting activities across various subreddits over a one-month period. Users and subreddits serve as nodes, while edges denote timestamped posting interactions. Each edge is enriched with a 172‑dimensional LIWC feature vector.
    \item \textbf{MOOC} is a bipartite interaction network derived from an online learning platform, where nodes correspond to students and course content units, such as videos and problem sets. Edges represent students’ access to specific content units, with each edge annotated by a 4‑dimensional feature vector capturing interaction characteristics.
    \item \textbf{Enron} is a large-scale dataset capturing email communications among employees of the ENRON energy corporation over a three‑year period. 
    \item \textbf{UCI} is an online communication network where nodes represent university scholars and edges correspond to messages posted by them. 
    \item \textbf{Can. Parl.} is a dynamic political network that captures the interactions among Canadian Members of Parliament (MPs) from 2006 to 2019. Each node corresponds to an MP representing an electoral district, and an edge is established when two MPs both vote “yes” on the same bill. The edge weight reflects the annual frequency with which one MP supports another through affirmative votes.
\end{itemize}   
The statistics for these datasets are reported in Table~\ref{tbl:dataset-stats}, where “\#N\&L Feat” denotes the dimensions of node and link features.

\subsection{Baselines.}
We select the following six popular baselines:
\begin{itemize}[leftmargin=1.6em]
    \item \textbf{TGN}~\cite{rossi2020temporal} is a representative memory-based model that maintains the evolving historical information of each node through a memory module and updates the memory upon the occurrence of new events. At the same time, it models dynamic graph neural networks through message functions, message aggregation, and memory update mechanisms.
    \item \textbf{CAWN}~\cite{wang2021inductive} first extracts multiple causal anonymous walks per node to capture the underlying causal patterns of network evolution and construct relative node identities. Each walk is then encoded using a recurrent neural network, and the encoded walks are subsequently aggregated to generate the final representation of the node.
    \item \textbf{TCL}~\cite{wang2021tcl} constructs each node’s interaction sequence by performing a breadth-first search over its temporal dependency subgraph. It then employs a graph transformer that jointly captures structural and temporal information to learn node representations. Furthermore, TCL introduces a cross-attention mechanism to model the interdependence between pairs of interacting nodes.
    \item \textbf{GraphMixer}~\cite{cong2023we} is a lightweight temporal graph learning model that combines a fixed time encoding function with an MLP-Mixer-based link encoder and a mean-pooling-based node encoder. It achieves strong performance in temporal link prediction with a simple, non-attention-based architecture.
    \item \textbf{DyGFormer}~\cite{yu2023towards} is a Transformer-based model for dynamic graph learning. It segments historical interactions into patches to capture long-term dependencies and uses neighbor co-occurrence encoding to model correlations between source and target nodes, enabling effective representation of both temporal and structural dynamics.
    \item \textbf{TPNet}~\cite{lu2024improving} leverages a temporal walk matrix projection with efficient random feature propagation to jointly capture temporal and structural dynamics for accurate and scalable dynamic link prediction.
\end{itemize}

\section{Implementation Details}
To ensure consistent performance comparisons, we adopt the implementations provided by DyGLib~\cite{yu2023towards}, a unified temporal graph learning library with well-tuned hyperparameters for each baseline. For baselines that are not included in DyGLib (e.g., TPNet~\cite{lu2024improving}), we use their official implementations. All models are trained using the Adam optimizer and supervised binary cross-entropy loss. Training is performed for up to 100 epochs with early stopping based on validation performance, using a patience of 20 epochs. The model achieving the best validation performance is selected for testing. For all datasets, we set the learning rate to 0.0001 and the batch size to 200. The dimensions of both node and edge features are fixed at 172 across all models, while the time encoding dimension is consistently set to 100. To ensure robustness, each method is executed five times with random seeds from 0 to 4, and the average results are reported.

We select two representative baselines as the backbones for our \textbf{Graph2Video}: the sequence‑based method DyGFormer~\cite{yu2023towards} and the temporal random walk method TPNet~\cite{lu2024improving}. We integrate our model into these two architectures and evaluate their performance on the dynamic link prediction task. Specifically, for Graph2Video, we train the models on different datasets using the best configurations of the two backbones. To enable the progressive incorporation of video features, we introduce a separate learning rate, denoted as \texttt{video\_lr}, dedicated to the video branch. In addition, a gradient scaling strategy is applied to video-related parameters to moderate their updates during training. This prevents the video features from dominating the optimization process in the early stages and facilitates smoother, more stable fusion with backbone representations. For hyperparameter tuning, \texttt{video\_lr} is selected from $\{10^{-4}, 10^{-5}, 10^{-6}, 10^{-7}\}$, and \texttt{grad\_scale} from $\{1, 10^{-5}, 10^{-6}, 10^{-7}, 10^{-8}, 10^{-9}\}$. During the fusion process, we also tune the coefficient $\alpha$, which controls the contribution of video feature representations, with candidate values drawn from $\{0.001, 0.005, 0.01, 0.02, 0.03\}$.
Specifically, in constructing the spatial scope of our Graph2Video, we set $k = 1$ and utilize the 1-hop temporal neighborhood of each endpoint. For the temporal scope, we partition the historical interval $[t_0, t^*]$ into $F = 16$ uniform segments to form temporal frames. Within each frame, a local subgraph is built by selecting the $s$ most recent 1-hop neighbors of each endpoint node. To assess the impact of spatial granularity, we vary $s \in \{8, 16, 32\}$ in our ablation study.
All experiments are conducted on a Linux server running Ubuntu, equipped with an Intel Xeon Silver 4210 CPU (2.20 GHz) and an NVIDIA Quadro RTX 8000 GPU with 48 GB of memory.

\section{Additional Experimental Results}
% We report the AUC results for transductive and inductive link prediction under three negative sampling strategies in Table~\ref{tab: AUC_trans} and Table~\ref{tab: AUC_ind}, respectively.

Table~\ref{tab: AUC_trans} and Table~\ref{tab: AUC_ind} present the AUC scores of DyGFormer+ and TPNet+ in comparison with state-of-the-art dynamic graph models on five real-world datasets. The evaluation is conducted under both transductive and inductive settings, with three negative-sampling strategies (random, historical, and inductive). For clarity, all numerical results are scaled by 100, and the best and second-best performances in each column are highlighted in \textbf{bold} and \underline{underline}, respectively.

The results show that integrating Graph2Video consistently improves the performance of both DyGFormer and TPNet across most datasets and sampling strategies. In particular, TPNet+ achieves the best overall performance in most cases, achieving the lowest average rank under both transductive and inductive settings. The improvements are especially noticeable under the more challenging historical and inductive negative sampling strategies, where modeling long-range temporal dependencies and complex interaction patterns becomes more difficult. These results indicate that the proposed Graph2Video framework effectively captures spatio-temporal structural dynamics and provides complementary information to existing dynamic graph encoders, leading to better link prediction performance.
% \bibliography{aaai2026}
% \end{document}

\end{document}